\documentclass[11pt]{article}

\usepackage[letterpaper,margin=1in]{geometry}

\usepackage[utf8]{inputenc} 
\usepackage[T1]{fontenc}\usepackage{lmodern}\usepackage{anyfontsize} 
\usepackage{natbib}         
\usepackage{url}            
\usepackage{booktabs}       
\usepackage{amsfonts}       
\usepackage{amsmath}        
\usepackage{nicefrac}       
\usepackage{microtype}      
\usepackage{xcolor}         
\usepackage{tabularx}
\usepackage{graphicx}       \usepackage[percent]{overpic} 
\usepackage{listings}       
\usepackage{float}          
\usepackage[colorlinks=true, citecolor=purple, linkcolor=black, urlcolor=blue]{hyperref}
\lstdefinestyle{eml}{
  basicstyle=\ttfamily\scriptsize,
  breaklines=true,
  frame=single,
  backgroundcolor=\color{gray!5},
  numbers=none,
  columns=fullflexible,
  keepspaces=true,
  showstringspaces=false,
}

\title{\textsc{CogGym}: Towards Large-Scale Comparative Evaluation of Human and Machine Cognition}
\author{\begin{minipage}{0.97\textwidth}\centering\setlength{\parindent}{0pt}\renewcommand{\textbf}[1]{\mbox{\normalfont ##1}}%
\fontsize{9}{10.8}\selectfont
\textbf{Lance Ying}$^{1,2}$ \quad
\textbf{Jinzhou Wu}$^{3}$ \quad
\textbf{Yingshan Susan Wang}$^{1}$ \quad
\textbf{Shivam Aarya}$^{4}$ \quad
\textbf{Luca M. Schulze Buschoff}$^{5}$ \quad
\textbf{Harry Chen}$^{6}$ \quad
\textbf{Katherine M.\ Collins}$^{1,7,8}$ \quad
\textbf{Andrea de Varda}$^{1}$ \quad
\textbf{Shuhao Fu}$^{9}$ \quad
\textbf{Sean Dae Houlihan}$^{10,15}$ \quad
\textbf{Akshay K. Jagadish}$^{8}$ \quad
\textbf{Guangyuan Jiang}$^{1}$ \quad
\textbf{Samuel Kiegeland}$^{11}$ \quad
\textbf{Tetsu Kurumisawa}$^{1}$ \quad
\textbf{Rongzhi Liu}$^{12}$ \quad
\textbf{Ryan Liu}$^{8}$ \quad
\textbf{Ningshan Ma}$^{1}$ \quad
\textbf{Kathryn McGregor}$^{8}$ \quad
\textbf{Younes Strittmatter}$^{8}$ \quad
\textbf{Polina Tsvilodub}$^{13}$ \quad
\textbf{Jacob Hoover Vigly}$^{14}$ \quad
\textbf{Sarah Wu}$^{15}$ \quad
\textbf{Enjie Xu}$^{16}$ \quad
\textbf{Yiling Yun}$^{16}$ \quad
\textbf{Kelsey Allen}$^{17,18,19}$ \quad
\textbf{Tyler Brooke-Wilson}$^{20}$ \quad\textbf{Brian Christian}$^{21}$ \quad\textbf{Evelina Fedorenko}$^{1}$ \quad
\textbf{Michael C. Frank}$^{15}$ \quad
\textbf{Michael Franke}$^{13}$ \quad
\textbf{Tao Gao}$^{16}$ \quad
\textbf{Samuel J.\ Gershman}$^{2}$ \quad
\textbf{Robert D. Hawkins}$^{15}$ \quad
\textbf{Jennifer Hu}$^{4}$ \quad\textbf{Julian Jara-Ettinger}$^{20}$ \quad\textbf{Max Kleiman-Weiner}$^{22}$ \quad\textbf{Sydney Levine}$^{23}$ \quad\textbf{Tal Linzen}$^{23}$ \quad
\textbf{Hongjing Lu}$^{16}$ \quad\textbf{Timothy J. O'Donnell}$^{19,24,25}$ \quad\textbf{Desmond C. Ong}$^{26}$ \quad\textbf{Steven T. Piantadosi}$^{21}$ \quad
\textbf{Rebecca Saxe}$^{1}$ \quad
\textbf{Eric Schulz}$^{5}$ \quad
\textbf{Tianmin Shu}$^{4}$ \quad\textbf{Felix A. Sosa}$^{27}$ \quad\textbf{Ilia Sucholutsky}$^{28}$ \quad\textbf{Tan Zhi-Xuan}$^{29,30}$ \quad
\textbf{Tomer Ullman}$^{2}$ \quad\textbf{Fei Xu}$^{21}$ \quad\textbf{Ilker Yildirim}$^{20}$ \quad\textbf{Jian-Qiao Zhu}$^{31}$ \quad
\textbf{Thomas L.\ Griffiths}$^{8}$ \quad
\textbf{Tobias Gerstenberg}$^{15}$ \quad
\textbf{Kevin Smith}$^{1,\dagger}$ \quad
\textbf{Joshua B.\ Tenenbaum}$^{1,\dagger}$
\\[5pt]
\fontsize{8.2}{9.8}\selectfont $^{\dagger}$Co-senior authors. \\[2.5pt]
\fontsize{8.0}{9.6}\selectfont\mbox{$^{1}$Massachusetts Institute of Technology} \hspace{0.6em}\mbox{$^{2}$Harvard University} \hspace{0.6em}\mbox{$^{3}$Cornell University} \hspace{0.6em}\mbox{$^{4}$Johns Hopkins University} \hspace{0.6em}\mbox{$^{5}$Helmholtz Munich} \hspace{0.6em}\mbox{$^{6}$Massachusetts General Hospital} \hspace{0.6em}\mbox{$^{7}$University of Cambridge} \hspace{0.6em}\mbox{$^{8}$Princeton University} \hspace{0.6em}\mbox{$^{9}$Santa Fe Institute} \hspace{0.6em}\mbox{$^{10}$Dartmouth College} \hspace{0.6em}\mbox{$^{11}$EPFL} \hspace{0.6em}\mbox{$^{12}$University of Chicago} \hspace{0.6em}\mbox{$^{13}$University of T\"{u}bingen} \hspace{0.6em}\mbox{$^{14}$CHI-FRO} \hspace{0.6em}\mbox{$^{15}$Stanford University} \hspace{0.6em}\mbox{$^{16}$University of California, Los Angeles} \hspace{0.6em}\mbox{$^{17}$University of British Columbia} \hspace{0.6em}\mbox{$^{18}$Vector Institute} \hspace{0.6em}\mbox{$^{19}$Canada CIFAR AI Chair} \hspace{0.6em}\mbox{$^{20}$Yale University} \hspace{0.6em}\mbox{$^{21}$University of California, Berkeley} \hspace{0.6em}\mbox{$^{22}$University of Washington} \hspace{0.6em}\mbox{$^{23}$New York University} \hspace{0.6em}\mbox{$^{24}$McGill University} \hspace{0.6em}\mbox{$^{25}$Mila--Quebec AI Institute} \hspace{0.6em}\mbox{$^{26}$The University of Texas at Austin} \hspace{0.6em}\mbox{$^{27}$Prior Computers} \hspace{0.6em}\mbox{$^{28}$Purdue University} \hspace{0.6em}\mbox{$^{29}$National University of Singapore} \hspace{0.6em}\mbox{$^{30}$A*STAR Institute of Advanced Intelligence and Computing} \hspace{0.6em}\mbox{$^{31}$The University of Hong Kong}\end{minipage}
}
\date{}

\begin{document}








\maketitle

\begin{abstract}
Understanding and modeling human intelligence are parallel goals shared by artificial intelligence (AI) and cognitive science. As AI systems grow increasingly capable, in what ways do model responses resemble human responses, and where do they systematically diverge? The sheer breadth and diversity of the tasks humans can perform and think about pose a challenge for scalable and rigorous comparison between humans and models. We introduce CogGym, a scalable, unified framework grounded in cognitive science for systematically comparing model and human behavior on matched experimental trials. CogGym uses a semi-automated, human-in-the-loop pipeline to standardize diverse experimental paradigms into a task-agnostic Experiment Markup Language (EML), enabling reproducible and faithful comparison at scale. For initial release, we curate and standardize 258 cognitive experiments from 100 papers focusing on human commonsense reasoning, and evaluate 50 large language models against human responses. We find a clear scaling trend where larger and more recent AI models better reproduce human judgments. Yet AI models' improvement on such common reasoning tasks is considerably slower than the gains observed on formal-reasoning benchmarks like math and coding, and model--human fit remains well below human splithalf reliability ($R^2 = 0.93$ on text, $0.95$ on image, and $0.92$ on video)  with the best models achieving $R^2 = 0.59$ on text, $0.58$ on image, and $0.43$ on video experiments. We intend for CogGym to provide a living evaluation framework that continually incorporates new cognitive science experiments to characterize where model behavior resembles human behavior, where it systematically diverges, and how those patterns change as models and experiments evolve.
\end{abstract}



\section{Introduction}

Since its founding, the discipline of Artificial Intelligence (AI) has often focused on formalizing the computations that underlie human capabilities for thinking and reasoning \citep{turing1950computing,newell1976computer,minsky1961steps}. In the intervening decades, there have been numerous attempts to build or use AI systems as models of human thinking \citep{marr1982vision,minsky1986society,rumelhart1986parallel,tenenbaum2011grow}, yet the field still lacks a standardized framework for measuring how well today's machines simulate human thought and behavior. The rapid advancement of large-scale neural architectures, most notably Large Language Models (LLMs) and their multimodal and reasoning successors \citep{Brown2020Language,openai2023gpt4}, has yielded systems capable of exhibiting remarkably human-like cognitive behavior across capabilities such as language generation and logical reasoning \citep{bubeck2023sparks,Wei2022Emergent}. These behavioral similarities have motivated researchers to treat frontier models as computational hypotheses about the human mind \citep{binz2023using}. 

The prospect of building human-like intelligence in machines also generated immense enthusiasm due to its transformative potential in science and other applications \citep{ying2025}. Highly accurate computational analogs of human cognition could serve as digital twins, enabling researchers to simulate behavioral phenomena at unprecedented scales and piloting experimental designs before deploying them in human populations~\citep{anthis2025position, ashokkumar2026large, park2022social}. From an applied perspective, human-like machines could accelerate product development and user testing by simulating diverse human reactions~\citep{simile2026cvs}. Recent research has also suggested that endowing machines with human-like cognitive architectures is one path towards safer, more effective, and more intuitive human-AI interaction~\citep{ho2022cognitive, collins2024building}.

However, despite these technological advances and potential benefits, the degree to which contemporary AI models behaviorally align with humans remains poorly understood and fiercely debated. Indeed, theorists of cognitive science have argued that fully replicating human behavior with an AI model is an inherently intractable problem \citep{van2024reclaiming,rich2021hard}. These debates raise a more precise question: in what ways do model responses resemble human responses, and where do they systematically diverge? The prevailing paradigm in the AI community relies on standardized benchmarks that aggregate performance against objective answer keys \citep{hendrycks2021measuring,Wang2018Glue,Wang2019Superglue,liang2023holistic} (e.g., answering standardized test questions or solving programmatic puzzles). Cognitive-science experiments instead use theoretically motivated manipulations to compare graded response patterns across items, conditions, and participants; objective benchmark accuracy rarely captures structural nuances of human behavior, such as systematic biases \citep{kahneman1979prospect}, graded probabilistic judgments \citep{griffiths2006optimal}, and the individual differences in responses across human participants \citep{wong2025,ying2025,tan2024devbench}.

A natural pathway to evaluate intelligence in machines is to test AI models on the already large and continually expanding space of diverse cognitive science experiments, the very paradigms that have been developed over decades to characterize human cognition. Indeed, within the cognitive sciences, there has been increasing interest in evaluating frontier models using classical psychological paradigms \citep{binz2023using,hagendorff2023human,dasgupta2022language,collins2022structured,ullman2023large,stojnic2023commonsense,frank2023baby, liu2024honesty}. However, these efforts largely remain fragmented: individual studies often focus on isolated topics or phenomena using idiosyncratic experimental setups, resulting in a scattered literature that lacks systematicity. The findings are also strikingly mixed: some studies report that models reproduce a wide range of human capabilities, biases, and classic experimental effects \citep{binz2023using,hagendorff2023human,cui2025replication,liu2025mind,zhu2024incoherent}, whereas others find that they diverge substantially from human behavior \citep{ullman2023large,mitchell2023comparing,schulzebuschoff2025visual,oh2026superhuman,wang2026simulating}, leaving the overall degree of alignment unresolved and the field polarized. In addition, the continuous release of new models, alongside the rapid updating of existing ones, means that studies examining a small set of models at a fixed point in time yield conclusions with rapidly diminishing relevance.

To address these limitations, recent efforts have begun to aggregate cognitive experiments into evaluation suites for LLMs. However, standardizing hundreds of heterogeneous cognitive experiments in a common executable format poses a substantial technical challenge. Such experiments are highly diverse in modality (text, image, video), are implemented with different software frameworks (e.g., PsychoPy \citep{peirce2007psychopy}, psiTurk \citep{gureckis2016psiturk}, jsPsych \citep{de2015jspsych}, MATLAB, and custom presentation platforms), and many of the original repositories are outdated, incomplete, or poorly documented. Recent attempts \citep{centaur2024,binz2023turning,hu2026simbench,lei2026humanllm, coda2024cogbench} rely on labor-intensive manual standardization that is difficult to scale, and cover only text-based tasks across a handful of topics, leaving the more complex image and video paradigms largely unaddressed. In addition, because such efforts are largely crowdsourced or scraped, they often translate studies into prompts without systematically checking whether instructions, stimuli, trial structure, and response formats match the original experiments or whether published human response patterns can be replicated. Mismatches introduced during prompt conversion can systematically distort model--human comparisons~\citep[e.g.,][]{wang2024large}. 

In this paper, we introduce \textbf{CogGym}, a scalable framework for systematically comparing AI systems with human behavior. Unlike a static benchmark, CogGym is reusable scientific infrastructure for sourcing, standardizing, administering, and re-running cognitive science experiments with current and future AI systems and new human participants. It builds a scalable, semi-automated pipeline that leverages AI agents with a human-in-the-loop to convert diverse cognitive experiments into a standardized Experiment Markup Language (EML), a high-level representational framework that abstracts the underlying logic of behavioral experiment paradigms (experimental design, stimuli, trial structure, and response type) into structured, standardized specifications. Crucially, because EML fully specifies an experiment, it can be used both to rerun studies on human participants---filling in missing data, increasing participant numbers, or replicating previously reported results---and to administer the same experiments to AI models. The common specification allows us to verify the standardized implementations against the original studies and new human data before asking where and how AI models diverge from humans.

To populate the platform, we sourced experiments from over 30 research labs specializing in computational modeling of human cognition, curating a diverse collection of 258 experiments from 100 published papers. We translated each experiment into EML using the CogGym pipeline and deployed this battery to evaluate 50 frontier and open-weight large language models. For an updated subset of 15 experiments, we compared human replication data and model responses against the original human data to assess data and experiment quality.

CogGym's coverage across experiments, topics, modalities, and response formats enables us to investigate fundamental questions about the behavioral alignment of AI systems:

\begin{enumerate}
    \item \textbf{Where do today's frontier AI models behaviorally align with humans, and where do they diverge?} How well do models match mean judgments, response distributions, and systematic patterns across experiments?
    \item \textbf{What model properties are associated with stronger behavioral alignment?} Do larger or reasoning-enabled models produce responses that more closely match human judgments?
    \item \textbf{On which tasks do models most and least align with humans?} Across topics and stimulus modalities, where do models most closely align with human behavior, and where do they diverge most dramatically?
\end{enumerate}

Our large-scale evaluation shows that cognitive alignment does increase with both model size and recency, but the slope is far more gradual than the steep, near-saturated gains these same models show on standardized formal-reasoning benchmarks in STEM~\citep{rein2023gpqa}, mathematics~\citep{hendrycksmath2021}, and coding~\citep{chen2021evaluating}. In addition, even the best aligned models still remain far from matching human behavioral responses, and model--human fit on video experiments remains markedly lower than on text and image experiments. 



\begin{figure}[t]
    \centering
    \includegraphics[width=\textwidth]{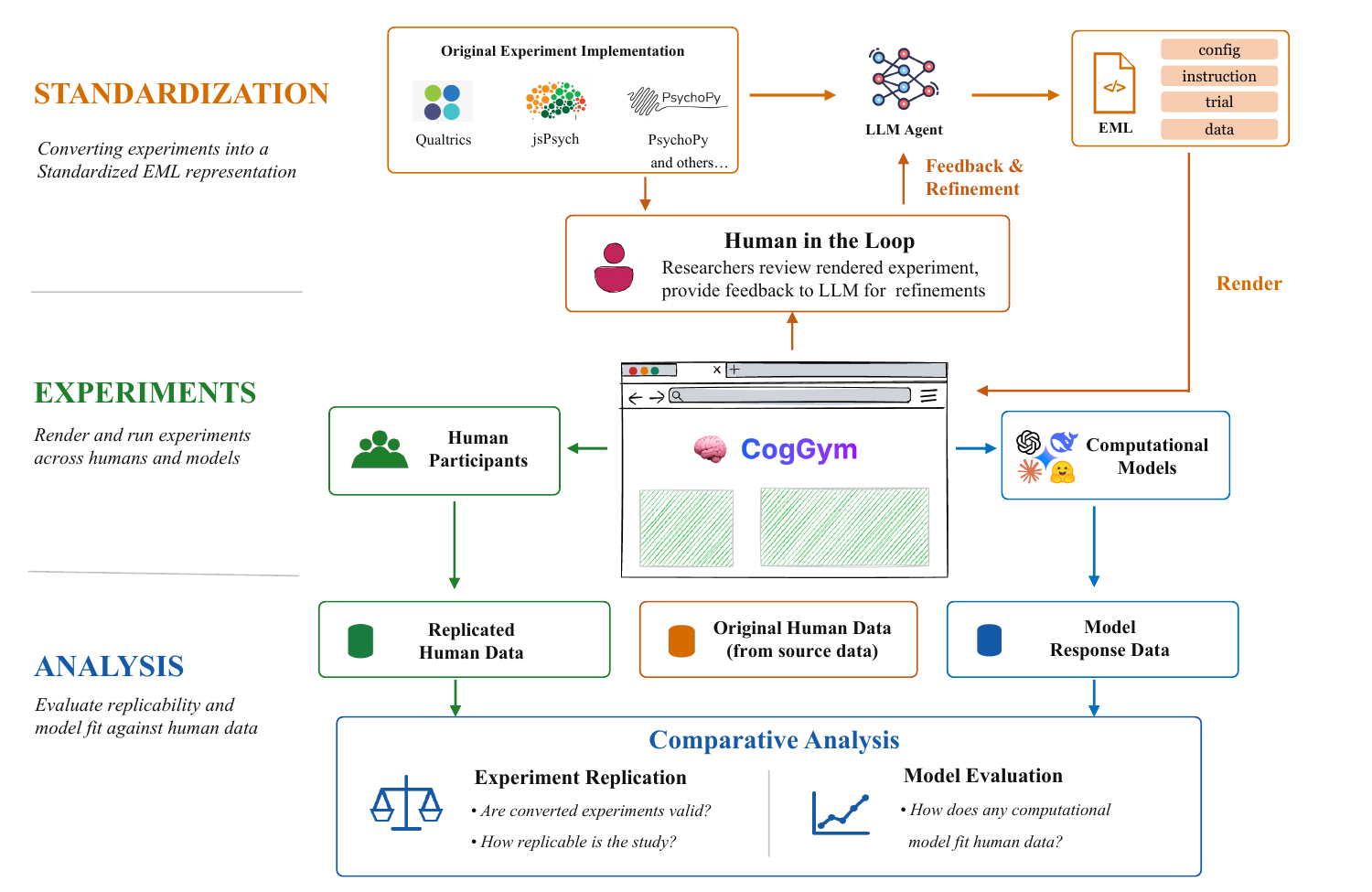}
    \caption{\textbf{CogGym workflow.} CogGym proceeds in three stages. First, original experiments implemented in diverse frameworks are converted into a standardized Experiment Markup Language (EML) representation through an AI agent with human-in-the-loop supervision. Second, the standardized experiments are rendered and administered to human participants and AI models. Lastly, human and model responses are analyzed jointly to evaluate experiment validity and model--human alignment.}
    \label{fig:eml_pipeline}
\end{figure}

\section{CogGym}

CogGym is a large-scale, scalable evaluation framework for comparing machine and human behavior. Instead of measuring whether models produce the objectively correct answer, CogGym examines whether models capture the behavior of human participants under the same experimental conditions. It leverages a scalable pipeline to standardize diverse cognitive experiments into a common Experiment Markup Language (EML), which can be administered to AI models as well as new online human participants to measure how agents reason with uncertainty, graded judgments, and response variability across a broad range of cognitive tasks. The experiment and dataset release, documentation, and information on contributing new experiments can be found at \url{https://coggym.org}.

\begin{figure}[t!]
    \centering
    \includegraphics[width=\textwidth]{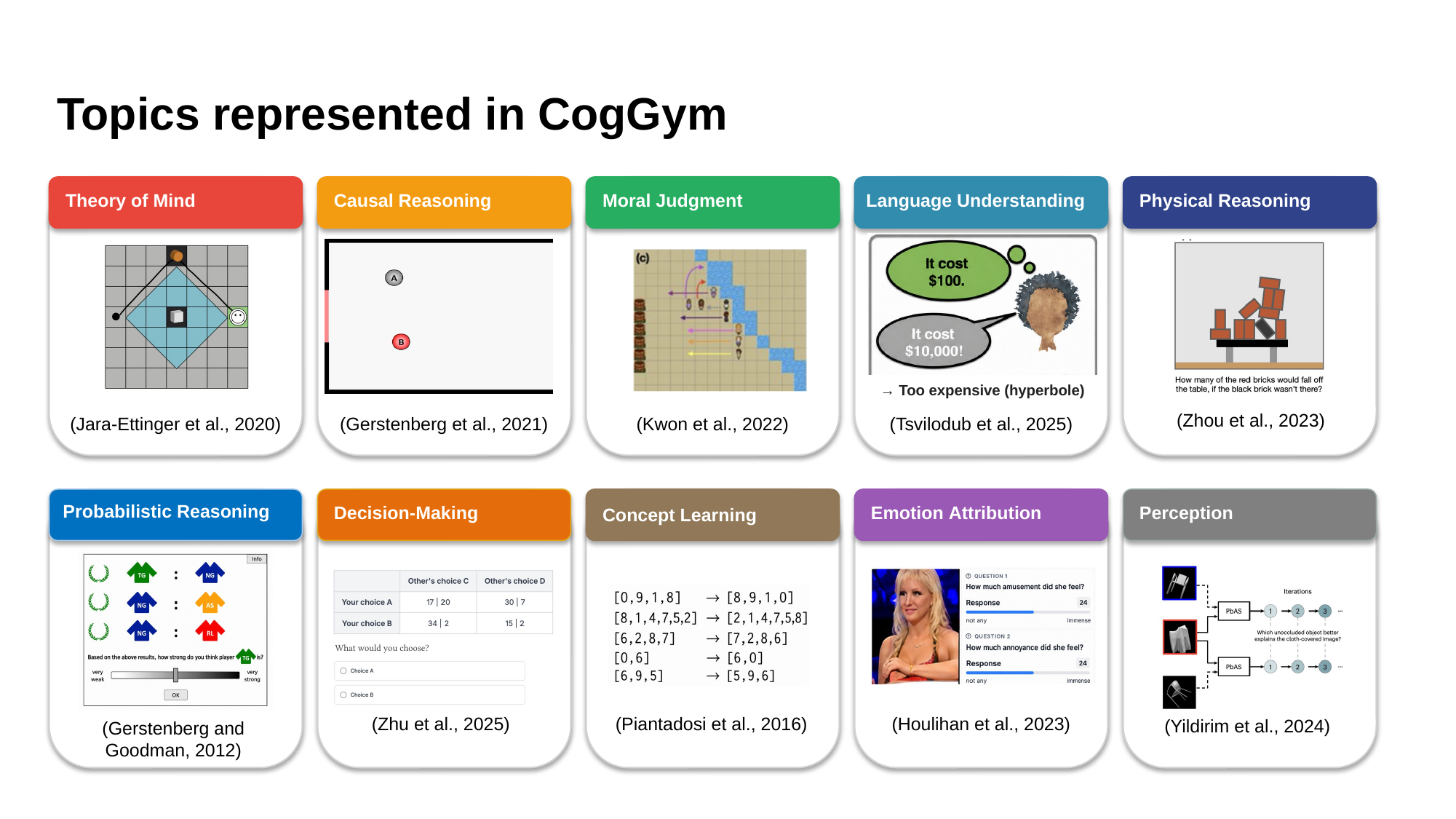}
    \caption{\textbf{Topics represented in CogGym.} CogGym experiments span a broad range of commonsense reasoning tasks, including physical reasoning, causal learning, concept learning, social cognition, moral judgment, probabilistic inference, pragmatic language use, and more. The studies were sourced from over 30 contributing research labs. These topics are derived from keywords reported in the source papers.}
    \label{fig:domains}
\end{figure}
\nocite{jaraettinger2020naive,gerstenberg2021counterfactual,tsvilodub2025nonliteral,piantadosi2016logical,zhu2025capturing,houlihan2023emotion,yildirim2024perception,zhou2023jenga,gerstenberg2012pingpong,kwon2022flexibility}

\subsection*{Sourcing Cognitive Science Experiments}

While CogGym supports comparisons between humans and machines across a wide range of tasks, we primarily focus on human commonsense reasoning: how people form and deploy intuitive theories of the physical and social world to solve everyday problems. A core feature of these tasks is that many of them have no objective ground truth, such as who is to blame when multiple people played a causal role in a bad outcome.

We collaborated with over \textbf{30 research labs} that study human commonsense reasoning to curate the first batch of experiments on CogGym, containing \textbf{258 experiments} from \textbf{100 published papers}. \autoref{fig:domains} summarizes the topics represented in the current experiment set, using keywords reported in the source papers, including theory of mind, causal reasoning, moral judgment, language understanding, physical reasoning, probabilistic reasoning, decision-making, concept learning, emotion attribution, and perception. The experiments feature text, images, and videos. The present evaluation analyzes response content and distributions; response times are outside its scope.

The CogGym workflow is divided into three parts---standardization, experiments, and analysis---which we illustrate in \autoref{fig:eml_pipeline}. We describe each stage below.

\subsection*{Experiment Standardization}

The first stage converts the original cognitive experiments into a common experiment representation. CogGym begins with source materials from an original study, including experiment code, stimuli, instructions, trial logic, and human behavioral data. These materials come from heterogeneous frameworks such as Qualtrics, jsPsych, PsychoPy, MATLAB, or custom JavaScript, making it impractical to use the raw source data to directly evaluate AI models.

To address this, we develop a unifying Experiment Markup Language (EML): a model-agnostic JSON representation of a behavioral experiment as executable data rather than framework-specific code. An EML experiment consists of several components: experiment configuration (e.g., randomization logic, and trial order), instructions (information and tutorial presented to the participants), trials (stimuli and queries), and human data. These components are all specified in a shared JSON format, allowing for a common specification that covers a wide range of standard cognitive science experiments, and making them easily interpretable to researchers. We provide an example in \autoref{fig:eml_example}. Detailed specifications and examples can be found in Appendix~\ref{sec:eml_details}. The EML can then be rendered to publish an experiment on the CogGym platform, which enables experiment visualization for researchers and human data collection.

\begin{figure}[t]
    \centering
    \includegraphics[width=\textwidth]{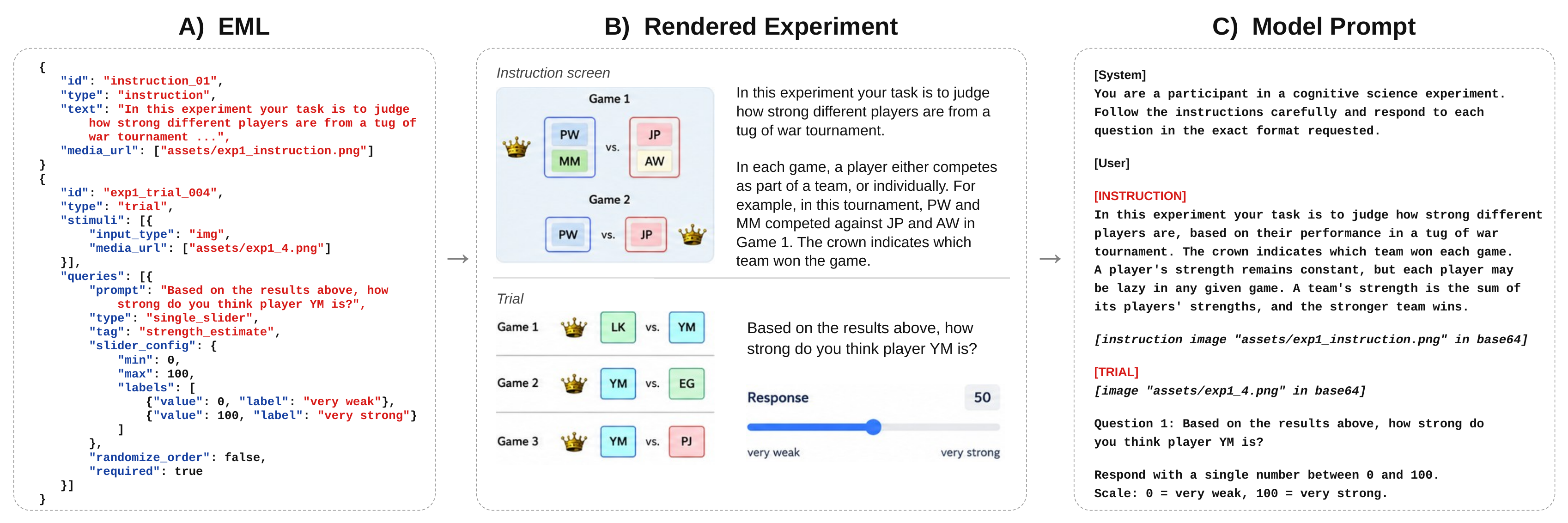}
    \caption{\textbf{Example of an EML experiment.} \textbf{A}: EML specification for the experiment, which encodes the instructions, trials (stimuli and response queries), and human data of the original study as executable data (adapted from \citealt{gerstenberg2012pingpong}). \textbf{B}: The same EML rendered on the CogGym platform as an interactive web experiment for human data collection. \textbf{C}: The matched model prompt automatically constructed from the same EML. Because the human task and the model prompt are generated from a single specification, participants and models receive closely matched instructions, stimuli, and response options, enabling a direct comparison of their responses.}
    \label{fig:eml_example}
\end{figure}

To convert experiments to EML, we developed an agentic AI workflow to produce an initial standardized specification. CogGym then renders the specification for a researcher to compare side by side with the original experiment. The researcher checks whether the stimuli, instructions, and response widgets display the same information and ask the same questions as the original paradigm. When discrepancies are found, the researcher provides feedback and the specification is revised by the AI agent. This generation and human-in-the-loop refinement process repeats until the CogGym experiment faithfully reproduces the original experiment as judged by the researcher. On average, converting an experiment into EML takes approximately one hour. Appendix~\ref{sec:experiment_checking} describes the EML generation and validation process in detail.

\subsection*{Data Collection from Models and Humans}

The second stage administers standardized experiments to humans and models for data collection. For model evaluation, the same EML trial objects are passed to a model harness. The harness converts each trial into a standardized prompt containing the experimental instructions, stimuli, and response options shown to human participants. Model outputs are parsed to the same query tags used for human responses, allowing a shared format for model and human data for analyses. 

For human validation or replication, CogGym renders each EML specification as a web experiment and records new human responses under the same trial and query identifiers used by the original dataset. This enables new human data to be compared directly against the source study, allowing the benchmark to test whether converted experiments preserve the intended behavioral signal.

\subsection*{Evaluating Model and Human Behavior}

The final stage places the original human data, newly replicated human data, and model responses into a common trial-level format, so all three are directly comparable. We then evaluate how closely each model matches human behavior: we quantify the degree of model--human alignment, visualize it experiment by experiment, and inspect qualitative examples to see where models align or diverge from human behavior. In parallel, we compare the original human data against the newly collected replication data to estimate the reliability of each experiment and its original data, excluding experiments that prove unreliable.

\section{Experiments}
\label{sec:experiments}

We evaluate 50 proprietary and open models spanning multiple model families, sizes, and release dates from April 2024 to April 2026. Each model is administered the full CogGym battery under matched conditions (default configuration with identical prompts), with 10 runs per model for each trial to obtain stable estimates. The prompt and harness for the model experiments are shown in Appendix~\ref{sec:model_eval_details}.

For each experiment, we compute $R^2$, the squared Pearson correlation between human and model mean responses.  For ordinal responses (e.g., slider ratings), each item contributes one data point: the model's mean response paired with the human mean. For categorical responses (e.g., multiple-choice), each option contributes a data point: the model's selection probability (proportion of runs selecting that option) paired with the human selection proportion. Within each experiment, we compute one $R^2$ for each answer type (defined by response regime and native scale), pooling conditions or constructs that share that type and scale; we average equally across answer types and then report the mean across experiments. Higher values indicate closer agreement between model and human mean responses. 

To measure whether the model reproduces the full distribution of responses across participants, we additionally compute a \textbf{normalized distribution divergence} metric. For each item, we take the distance between the model and human response distributions---Earth Mover's Distance (EMD) for continuous items (normalized by the scale range) and Jensen--Shannon Divergence (JSD) for categorical items---rescaled to $[0,1]$. $0$ indicates the model matches exactly the human distribution. Because repeatedly sampled model responses are sharply peaked and show little variation (see Appendix~\ref{sec:peaked}), we follow \citet{meister-etal-2025-benchmarking} and prompt models to provide a verbalized distribution over response options to compare against the human response distribution. The prompts for querying the AI models are provided in Appendix~\ref{sec:model_eval_details}.

To obtain human reliability references for the two metrics, we estimate human split-half reliability by randomly splitting participants into halves and computing the metrics, averaged over 100 splits. We applied the Spearman--Brown correction to estimate full-sample reliability. The human split-half is computed over 179 of the 258 experiments (80 text, 37 image, 62 video; see Appendix~\ref{sec:splithalf_subset}), where individual-level response data exists and there is sufficient coverage (at least 10 data points per judgment). 

\section{Results}\begin{figure}[tp]
    \centering
    
    \includegraphics[width=\textwidth]{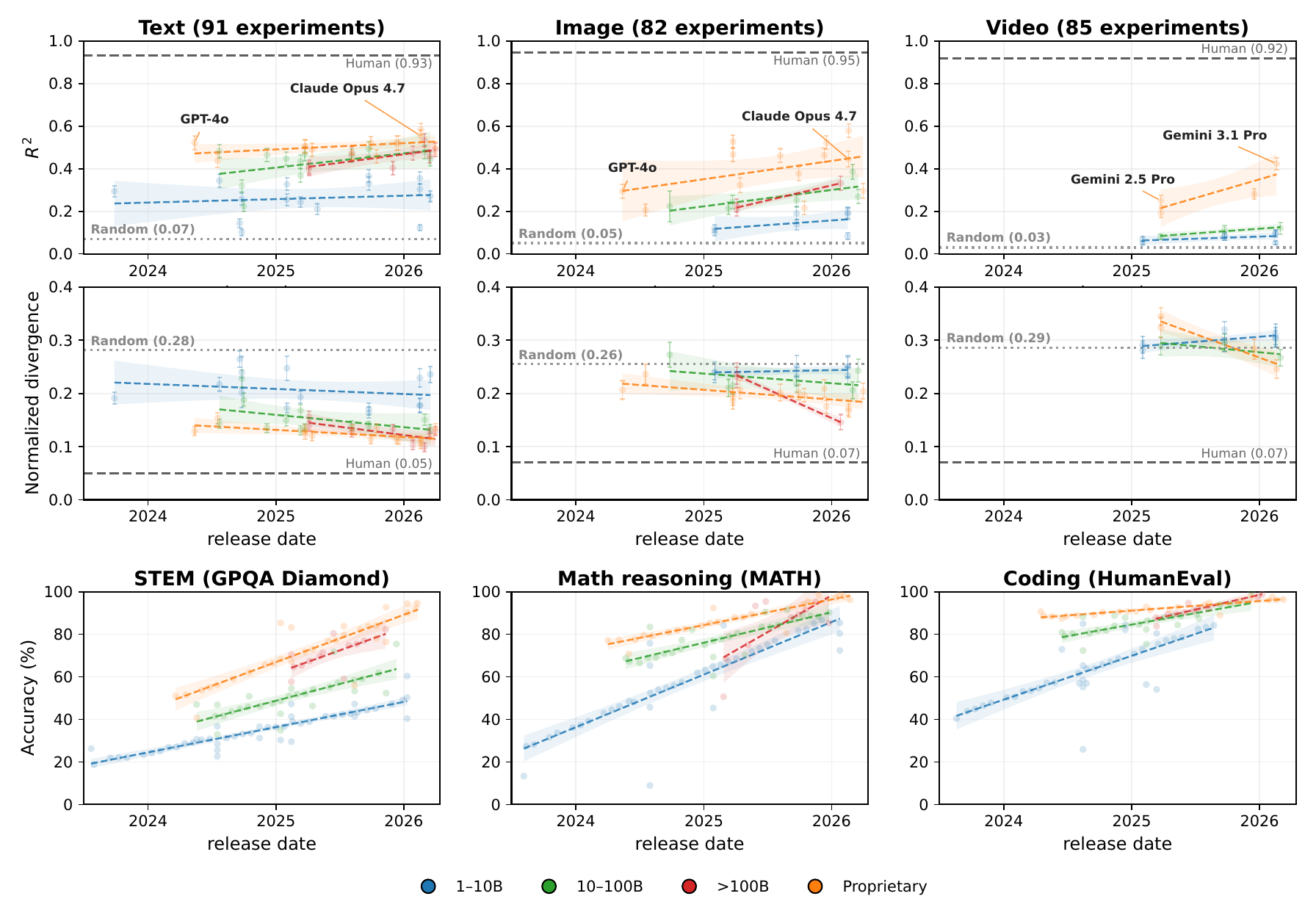}

    \caption{\textbf{Model performance on CogGym tasks and formal reasoning benchmarks plotted by model release date.} Models are color-coded by size, and selected models are labeled. \textbf{Top:} model--human fit $R^2$ across experimental tasks. \textbf{Middle:} normalized distributional divergence between AI models and humans. \textbf{Bottom:} AI models' performance on GPQA Diamond, MATH, and HumanEval performance. The plots show that models' improvement on formal reasoning benchmarks is much steeper than models' improvement on their alignment with humans on commonsense reasoning tasks. Error bars show $\pm 1$ standard error across experiments.}
    \label{fig:gpqa_marquee}
\end{figure}

\begin{figure}[ht!]
    \centering
    \includegraphics[width=\textwidth]{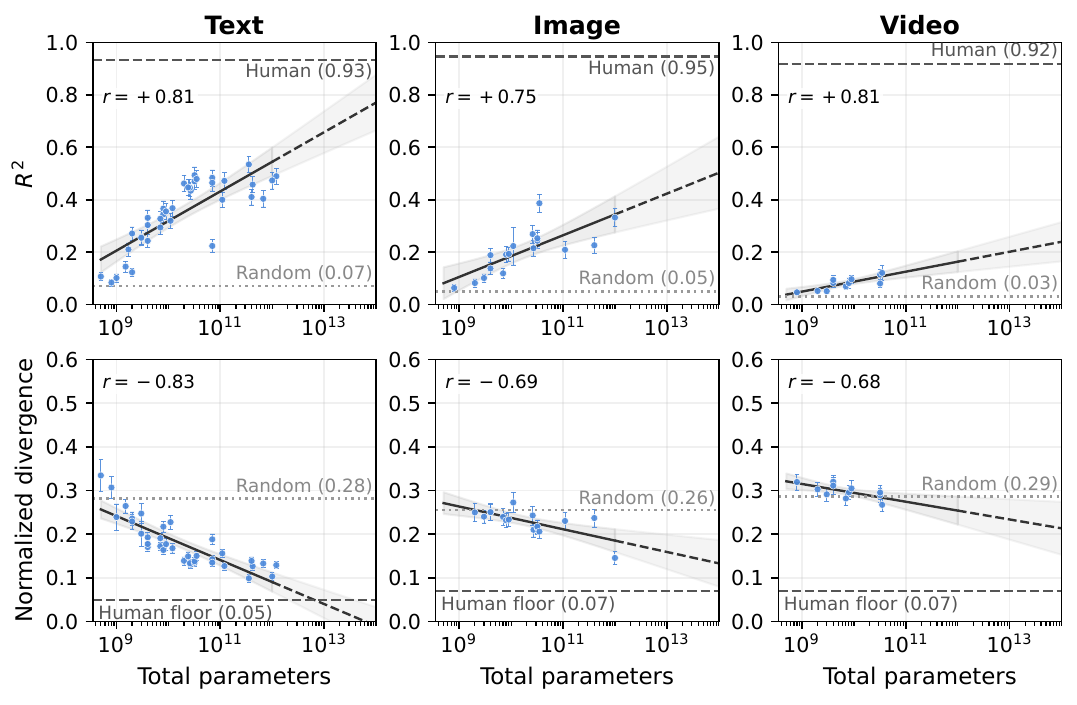}
    \caption{\textbf{Model--human alignment as a function of model size among open-weight models.} Each point is one open-weight model. \textbf{Top row:} model--human fit $R^2$ ($\uparrow$) rises with scale in every modality (text $r=+0.81$, image $+0.75$, video $+0.81$). \textbf{Bottom row:} normalized distribution divergence falls with scale in every modality (text $r=-0.83$, image $-0.69$, video $-0.68$). Error bars are per-model $\pm 1$ standard error. Dashed extensions show log-linear extrapolations. Shaded regions are 95\% confidence bands, shown more lightly over the extrapolated range.}
        \label{fig:scaling_r2}
\end{figure}

Our results suggest that  
larger, more recently released models show higher behavioral alignment with humans. \autoref{fig:gpqa_marquee} shows that model--human fit on CogGym has risen substantially over the past two years. On text and image experiments, the strongest models now achieve relatively high mean-response fit (best $R^2 = 0.59$ and $0.58$, respectively), although both remain below Spearman--Brown-corrected human reference $R^2$ values ($0.93$ and $0.95$). Their normalized distributional divergence has also narrowed, reaching $0.10$ on text and $0.15$ on image, compared with human floors of $0.05$ and $0.07$, respectively. Video tasks remain substantially more challenging (best $R^2 = 0.43$ vs. a corrected human reference of $0.92$; divergence $0.25$ vs.\ a human floor of $0.07$). Moreover, this progress has been considerably slower than the near-saturation gains observed on formal-reasoning benchmarks such as STEM (GPQA Diamond), math (MATH), and coding (HumanEval) over the same period (\autoref{fig:gpqa_marquee}).


\autoref{fig:scaling_r2} plots per-model $R^2$ against total parameter count for open-weight models (per-model $R^2$ and normalized-divergence scores by modality are reported in \autoref{tab:results} in Appendix~\ref{sec:full_results}). Model--human fit $R^2$ rises with scale in every modality (text $r = +0.81$, image $r = +0.73$, video $r = +0.81$), although fewer open-weight image and video models have valid scores. The same trend holds for distributional divergence (\autoref{fig:scaling_r2}, bottom row): divergence falls with scale in all three modalities (text $r = -0.84$, image $r = -0.79$, video $r = -0.68$), with all remaining well above the human split-half floor. 


While size and release date are positively correlated with alignment, they are also correlated with one another (larger models tend to be more recent). 
In order to estimate the independent effects of size and time, we applied a multivariable regression
$$y = \beta_{\text{size}} \log_{10}(N_{\text{params}}) + \beta_{\text{time}}\,d + \epsilon,$$ over the open-weight models in \autoref{tab:results}. Across the open-weight models, the scale coefficient $\beta_{\text{size}}$ is positive for $R^2$ in text, image, and video (all $p < .001$), and negative for distributional divergence (text $p < .001$, image $p = .004$, video $p = .036$). After controlling for size, release date is not significantly associated with text $R^2$ ($\hat{\beta}_{\text{time}} = +0.010$/yr, $p = .637$) or image $R^2$ ($+0.043$/yr, $p = .205$), but is positively associated with video $R^2$ ($+0.026$/yr, $p = .012$). Release date is not significantly associated with distributional divergence in text ($-0.006$/yr, $p = .501$), image ($+0.006$/yr, $p = .807$), or video ($+0.008$/yr, $p = .396$). Taken together, these results suggest that release date is not consistently associated with model--human fit once model size is taken into account among the open-weight models.

\begin{figure}[!t]
    \centering
    \includegraphics[width=\textwidth]{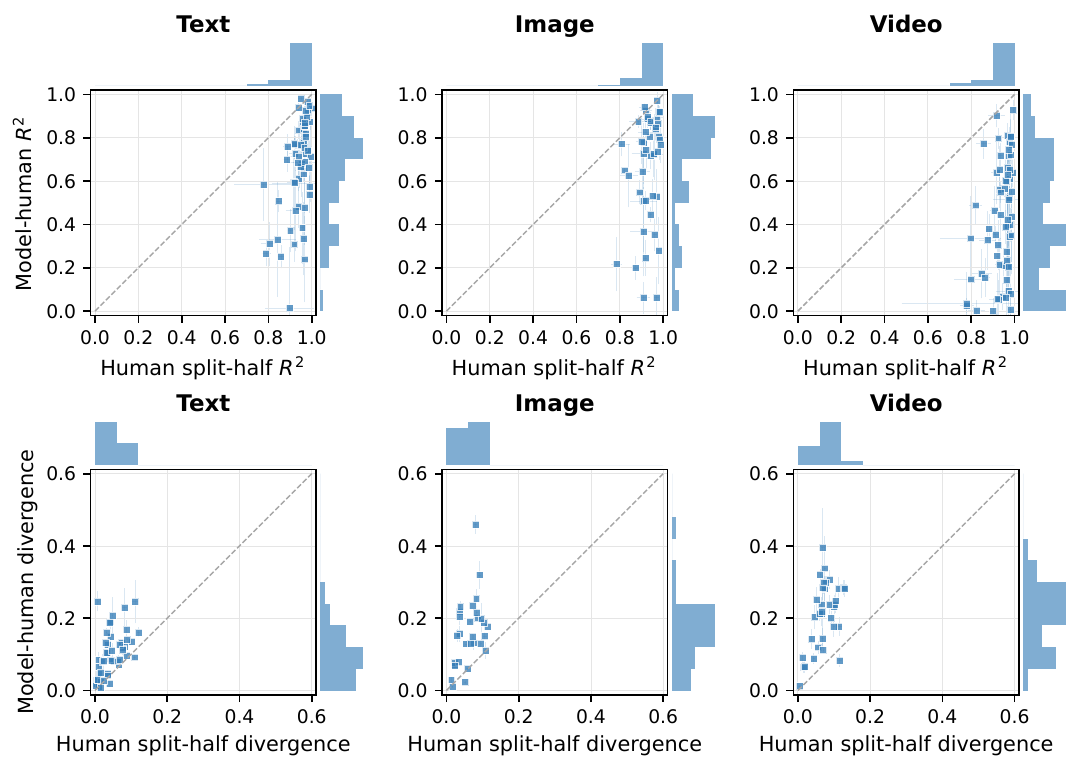}
    \caption{\textbf{Gemini 3.1 Pro vs.\ human split-half, per experiment.} Each point is one experiment. \textbf{(Top) $R^2$:} Human split-half $R^2$ vs.\ model--human fit $R^2$; $94\%$ of experiments fall below the parity diagonal. \textbf{(Bottom) Distributional Divergence}: human split-half normalized divergence (reliability floor) vs.\ model--human normalized divergence; Error bars in both panels are $\pm 1$ SE from 1,000-sample item bootstraps.}
    \label{fig:splithalf_scatter}
    \label{fig:splithalf_scatter_div}
\end{figure}

\begin{figure}[!htbp]
    \centering
    \includegraphics[width=\textwidth]{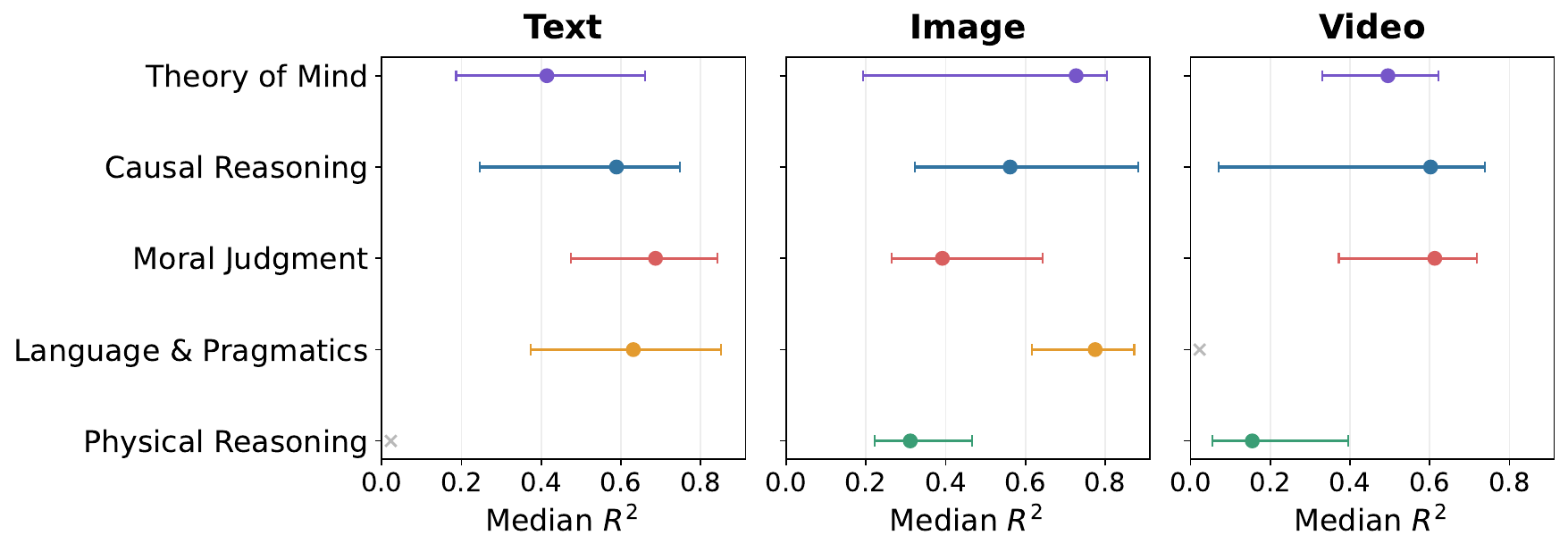}
    \caption{\textbf{Gemini 3.1 Pro's model--human $R^2$ by topic and modality.} Points show the median human--model fit across experiments; horizontal bars span the interquartile range (25th--75th percentiles). A cross marks a topic which lacks sufficient number of experiments. The plots show that models tend to align less with humans on physical reasoning tasks than other topics.}
    \label{fig:domain_per_model}
\end{figure}

\subsection*{Model cognitive alignment varies among experiment modality}

Across the full model set, mean per-experiment explained variance decreases from text ($R^2 = .39$) to image ($.29$) to video ($.14$). A crossed mixed-effects analysis of the 15 model groups evaluated on all three modalities, with random intercepts for model, study, and experiment and model-specific modality contrasts, confirmed all pairwise differences (text minus image: $\Delta R^2 = .13$, 95\% CI $[.05,.21]$; text minus video: $.26$, $[.17,.34]$; image minus video: $.13$, $[.05,.21]$; all Holm-adjusted $p < .002$). These are adjusted differences in model--human $R^2$, rather than differences between the pooled means. The same ordering held in an analysis of the full model set, although the best current model performs similarly on text and image ($R^2 = .59$ and $.58$, respectively). \autoref{fig:splithalf_scatter} plots Gemini 3.1 Pro fit against human split-half reliability. The model--human $R^2$ remains below this conservative human reliability reference for most experiments. In both text and image, most experiments are far from zero and in many experiments models fit moderately to strongly with humans. Video is much worse, with most experiments clustered between $0$ and $0.3$.

These differences may reflect perceptual limitations, context-length strain as video experiments tend to require longer context and more tokens, or the greater demands of tracking agents and objects in dynamic scenes.

\subsection*{Model fit varies within and across topics}

We use keywords reported in the 100 source papers to produce five common topic groups---Theory of Mind, Causal Reasoning, Moral Judgment, Language \& Pragmatics, and Physical Reasoning---and summarize model performance within and across these groups. \autoref{fig:domain_per_model} shows the median and interquartile range of Gemini 3.1 Pro's experiment-level $R^2$ by topic and modality. For Gemini 3.1 Pro, Physical Reasoning is the least aligned topic, with the lowest mean $R^2$ in both image ($.33$) and video ($.24$) experiments. 

However, within individual topics, the same model can fit some experiments nearly perfectly and others near chance. Large differences in model--human fit can even be observed across experiments within a single paper. Thus, experiment-level design and structure would be more predictive of alignment than broad topic labels.

\subsection*{Is alignment low because models are superhuman?} 

A natural concern is that divergence from human responses can reflect that models are more rational or competent than people. For example, many models today perform coding or math tasks better than average human participants~\citep{quan2025codeelo, elkishky2025competitive}. We argue against this interpretation based on the following observations.

\begin{figure}[t]
    \centering
    \includegraphics[width=\textwidth]{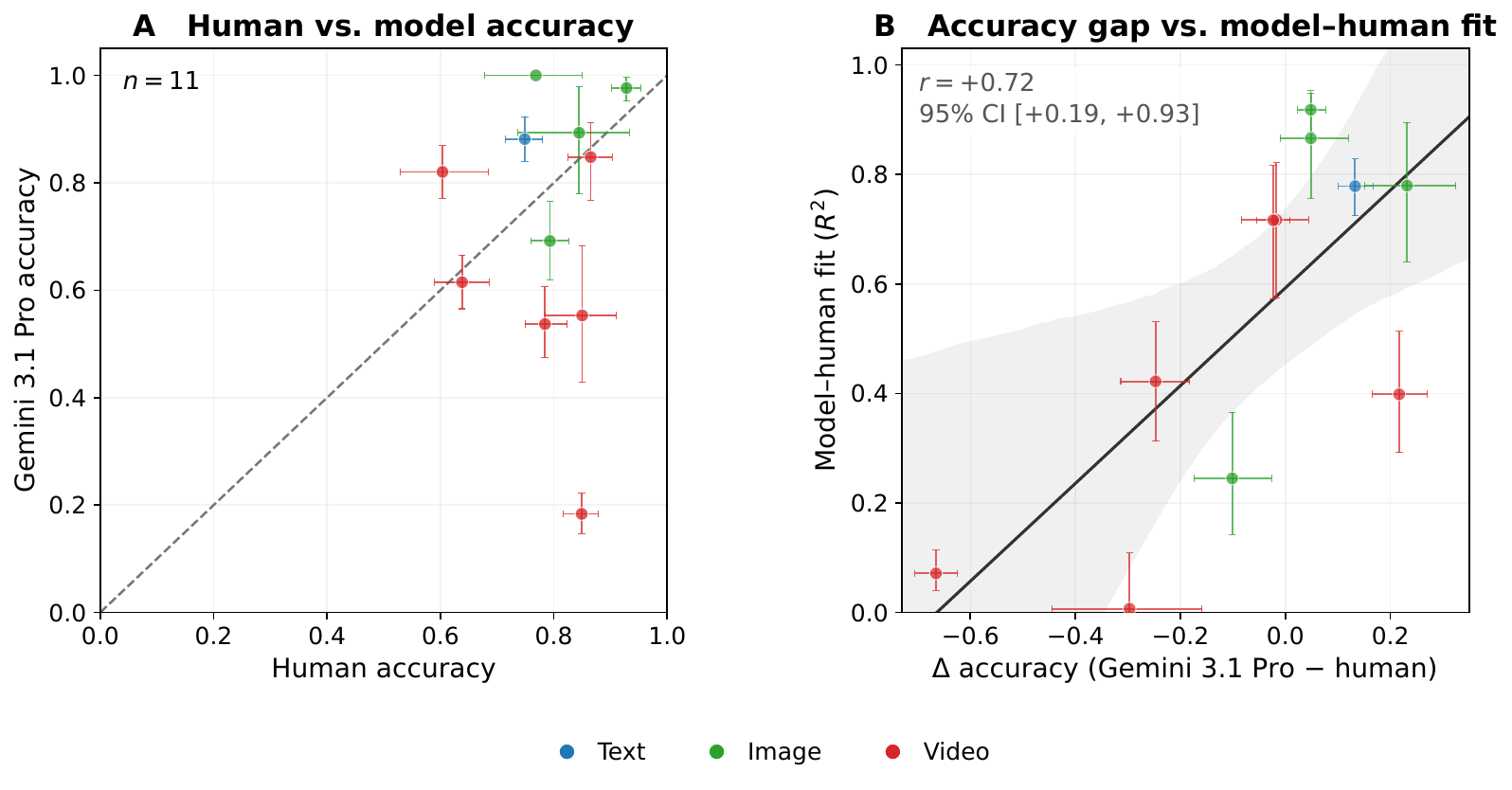}
    \caption{\textbf{Accuracy and model--human fit for Gemini 3.1 Pro vs Human participants.} Each data point indicates one experiment. (A) Mean human versus model accuracy for the 11 experiments that include accuracy metrics. The dashed diagonal marks equal accuracy ($y=x$); points above it favor the model and points below it favor humans. Overall humans participants achieve higher mean accuracy than Gemini 3.1 Pro. (B) Differences between model and human accuracy versus model--human $R^2$ on the same experiments, which shows a positive correlation between Gemini 3.1 Pro performance relative to humans' and model--human fit.}
    \label{fig:gt_acc}
\end{figure}

Firstly, we intentionally sourced experimenets where human performance is close to optimality: among the 258 experiments, only 11 experiments have reported accuracy metrics in their paper. On this subset of experiments, mean human accuracy ($0.79$) is not statistically different from Gemini 3.1 Pro accuracy ($0.73$; paired $t$-test, $p = .46$; \autoref{fig:gt_acc}). Within this subset, Gemini's accuracy relative to humans' is positively associated with its fit to human mean responses ($r = .72$, 95\% bootstrap CI $[.19,.93]$; \autoref{fig:gt_acc}). Thus, we don't observe any evidence for an inverse scaling law where superhuman models are less human-like.

Secondly, the rest of the experiments often elicit inherently subjective or normative judgments---how much responsibility or blame an agent deserves, how much one would pay someone to perform a task, whether an action is morally permissible, or how to interpret a sentence. For such questions there is no ground-truth answer key; we believe studying human-likeness on these domains is particularly valuable.

\subsection*{Evaluating the replicability and validity of the experiments}

Another concern is that low alignment could reflect noisy human data or errors introduced when translating an experiment into EML, rather than genuine model--human differences. We randomly sampled a subset of 15 experiments and recruited 521 human participants on Prolific to complete the experiments on CogGym and then compare human replication data and Gemini~3.1 Pro responses against the original human data (See Appendix~\ref{sec:replication_bound}). Across experiments, replication means agree more closely with the original human means (mean $R^2 = 0.84$) than do Gemini~3.1 Pro responses (mean $R^2 = 0.62$). These results indicate that the low $R^2$ observed between models and humans in these experiments is likely due to genuine model misalignment rather than experiment or data quality.

\subsection*{Contamination analysis}

To assess possible dataset contamination of the source studies in our evaluation, we asked AI models to identify the paper titles of published experiments from experimental materials without access to the Internet. Gemini 3.1 Pro achieved the highest exact-title recognition rate (29/175, 16.57\%), followed by Gemini 3 Flash (18/175, 10.29\%) and Claude Opus 4.7 (5/175, 2.86\%). Open-weight models exhibited lower mean exact-title recognition rates than proprietary models in this sample (0.11\% versus 3.26\%).

We next examined whether normalized Levenshtein similarity between the recalled and correct paper titles was associated with model--human fit ($R^2$). We find that the correlation is not significant for Gemini 3.1 Pro ($r=0.108$, $p=0.184$), Gemini 3 Flash ($r=0.122$, $p=0.111$), or Claude Opus 4.7 ($r=0.101$, $p=0.227$). These findings suggest that some models can recognize source studies, but recognition alone does not seem to affect the model's fit to human data.

\section{Discussion}

In this paper we propose CogGym, which establishes a reusable, executable infrastructure for comparing language and multimodal models with human behavioral data across a large collection of cognitive science experiments. Using this infrastructure, we find that model--human behavioral alignment remains well below human split-half reliability across a wide range of commonsense reasoning experiments, even as models have become sharply more capable on formal benchmarks such as math, coding, and STEM.

Why does behavioral alignment scale more slowly than performance on formal benchmarks? One plausible explanation is that contemporary training pipelines are particularly well matched to domains with verifiable answers. Mathematics and coding provide abundant automatically generated examples, exact correctness signals, and opportunities for reinforcement learning, test-time search, and automatic verification. CogGym experiments instead require models to reproduce human graded judgments in inherently subjective domains such as moral reasoning. Such tasks may be under-represented in the model training process. Although domains such as physical reasoning can have objective labels, they also require models to infer physical properties and dynamics from images or videos, rather than manipulate explicitly specified symbolic inputs. Physical reasoning tasks in CogGym often also involve more complex, continuous visual scenes, whereas tasks in other domains more commonly use static images or discrete, gridworld-like representations. This difference may contribute to the lower model--human alignment observed in physical reasoning, alongside challenges in reasoning about physical dynamics. Recent physics-oriented visual question answering benchmarks, including PhysBench and PhyX, likewise report substantial limitations in vision-language models' physical understanding \citep{chow2025physbench,shen2025phyx}.


While CogGym provides a scalable platform for comparing model and human behavior, it has several limitations, each pointing to a direction for future work:

\paragraph{Continual expansion of experiment set.}
Our initial 258 experiments are a modest start towards building a large-scale evaluation suite for comparing model and human cognition. As a living benchmark, CogGym will continually incorporate new experiments from contributing labs and newly published studies, thus building towards a more comprehensive evaluation platform that covers all aspects of cognition. Future work can also integrate automated experiment design \citep{jagadish2026closing,prystawski2026auto} to further extend the experiment suite, especially on topics where we expect to see greater divergence between AI models and humans. 

\paragraph{Deeper analysis.}

Many of the experiments in CogGym were designed with carefully controlled factorial manipulations, parametric variations, and theoretically motivated condition contrasts that enable far more fine-grained diagnoses of which specific aspects of cognition a model captures or fails to capture. For example, a social cognition experiment might reveal that a model correctly infers goals but not costs, or a causal reasoning task might show sensitivity to a mechanism but not counterfactual structure. Extracting these insights at scale requires more sophisticated automated analysis pipelines than the summary statistics we report here. We expect that future work on CogGym will develop tools for deeper, experiment-specific insights into the cognitive profile of AI systems and humans. Such analyses could move beyond asking ``how human-like is this model?''~toward more scientifically productive questions such as ``in what ways is this model human-like, and in what ways is it not?'' 

\paragraph{Interactive experiments.}The current EML specification supports single-turn, non-interactive paradigms, including one-shot economic games, but does not yet support repeated economic games, multi-agent social interactions, or sequential decision-making tasks where the stimulus depends on the participant's previous responses. 
Future work will extend the EML to account for richer task specifications, in order to test the performance of AI models against the wider universe of cognitive experiments.

\,

\noindent Going forward, we aim to grow CogGym as living, versioned scientific infrastructure that incorporates increasingly diverse and complex experiments, supports fresh human replications, and permits the same studies to be rerun as models change. We hope that it provides a scalable framework for identifying where model behavior resembles human behavior, where it diverges, and which experimental or model features  explain those patterns.

\section*{Acknowledgements}
We are grateful to Verona Teo, Lara Kirfel, Xinyi Lu, and Chuqi Hu for their help in validating the implementations of their experiments and ensuring that they faithfully reflected the original studies. We thank Jacob Andreas for his thoughtful and constructive feedback on the manuscript, and Charles Kemp for recommending studies from his lab for inclusion in CogGym. We also thank Ryan Truong and Raphael Yamamoto for their valuable assistance throughout the project. This work was supported by the MIT Quest for Intelligence.

\bibliographystyle{unsrtnat}
\bibliography{bibliography}

\clearpage
\appendix
\section*{Appendix}

\renewcommand{\thesubsection}{\Alph{section}.\arabic{subsection}}
\section{Experiment Specification and Validation}
\setcounter{subsection}{0}
\subsection{Experiment Markup Language (EML)}
\label{sec:eml_details}

EML is the interchange format that lets CogGym render heterogeneous cognitive experiments and administer the same trial logic to humans and models. The key design choice is to represent each experiment as data: flow, stimuli, response widgets, and human responses are explicit JSON objects with shared identifiers. This makes the experiment portable across web rendering, model prompting, and scoring.

\paragraph{Directory layout.} A complete EML experiment lives in a study subdirectory such as \texttt{StudyName/exp1/}:
\begin{lstlisting}[style=eml]
StudyName/
  README.md
  exp1/
    README.md
    config.json
    instruction.jsonl
    trial.jsonl
    human_data_mean.json
    human_data_ind.json
    assets/
      stimulus_001.png
\end{lstlisting}

\paragraph{1. Experiment metadata and flow.} The \texttt{config.json} file defines citation metadata, response types, and the sequence of modules and trials. The \texttt{experimentFlow} field is the bridge between the experiment-level design and the line-oriented instruction/trial files: every ID listed in \texttt{blocks} must appear in either \texttt{instruction.jsonl} or \texttt{trial.jsonl}.
\begin{lstlisting}[style=eml]
{
  "experimentName": "Example causal judgment task",
  "description": "Participants view a scene and rate how much one event caused another.",
  "paperDOI": "https://doi.org/10.1234/example",
  "taskType": ["Causal Reasoning"],
  "responseType": ["single-slider", "multi-choice"],
  "contributors": ["CogGym Team"],
  "stimuli_count": 1,
  "experimentFlow": [
    {
      "experimental_condition": "standard",
      "blocks": [
        ["instruction_01", "quiz_01"],
        ["trial_001"]
      ]
    }
  ]
}
\end{lstlisting}

\paragraph{2. Instructions and checks.} The \texttt{instruction.jsonl} file contains one JSON object per line. Modules can be plain instructions, practice trials, or comprehension quizzes. These modules are referenced by ID from \texttt{config.json}.
\begin{lstlisting}[style=eml]
{"id":"instruction_01","type":"instruction","text":"<p>You will see a scene and answer questions about what caused the outcome.</p>"}
{"id":"quiz_01","type":"comprehension_quiz","text":"Please confirm the task.","queries":[{"prompt":"What should you judge?","type":"multi-choice","tag":"quiz_task","option":["Color","Cause","Memory"],"answer":1}]}
\end{lstlisting}

\paragraph{3. Trials.} The \texttt{trial.jsonl} file also contains one JSON object per line. Each trial has an \texttt{id}, a \texttt{stimuli} array describing what is shown, and a \texttt{queries} array describing what responses are collected. Query \texttt{tag} values become the keys used for scoring.
\begin{lstlisting}[style=eml]
{"id":"trial_001","stimuli":[{"input_type":"img","media_url":["assets/stimulus_001.png"]},{"input_type":"text","text":"A ball hits a switch, and the machine turns on."}],"queries":[{"prompt":"How much did the ball cause the machine to turn on?","type":"single-slider","tag":"cause_rating","slider_config":{"min":0,"max":100,"default_value":50,"labels":[{"value":0,"label":"Not at all"},{"value":100,"label":"Completely"}],"show_label_values":true},"required":true},{"prompt":"Which event was most responsible?","type":"multi-choice","tag":"cause_choice","option":["The ball","The switch","The machine"],"randomize_order":false,"required":true}]}
\end{lstlisting}

EML supports \texttt{text}, \texttt{img}, and \texttt{video} stimuli. It supports the response formats needed for the benchmark: \texttt{single-slider}, \texttt{multi-slider}, \texttt{multi-choice}, \texttt{multi-select}, \texttt{ranking}, and free-text \texttt{textbox} responses. A \texttt{text-instruction} query can also insert explanatory text inside a trial without collecting a response.

\paragraph{4. Human responses.} Human data files use the same trial IDs and query tags. The mean file stores aggregate responses used for correlation, while the individual file stores participant-level responses used for split-half reliability.
\begin{lstlisting}[style=eml]
{
  "participants_info": {"count": 40, "age": 31.2, "gender": {"woman": 21, "man": 18, "other": 1}},
  "trial_001": {
    "cause_rating": 72.4,
    "cause_choice": {"The ball": 0.70, "The switch": 0.25, "The machine": 0.05}
  }
}
\end{lstlisting}

\paragraph{How EML executes.} CogGym executes EML in four steps. First, the renderer reads \texttt{config.json} and samples an experimental condition if multiple flows are present. Second, it resolves each block into instruction, quiz, practice, and trial objects using their IDs. Third, it renders each trial by placing the \texttt{stimuli} on the stimulus side of the interface and the \texttt{queries} as response widgets with the specified options or scale bounds. Fourth, the model harness converts the same trial object into a standardized prompt and parses the model response back into the query tags. Because the rendered experiment, model prompt, and human data all share the same trial IDs and tags, model and human responses can be aligned without experiment-specific scoring code.

\subsection{EML Generation and Validation}
\label{sec:experiment_checking}

To generate EMLs, we first construct Meta-EML (MEML), a specification layer for generating experiments in EML. A MEML specification describes an experiment's conditions, trial structure, randomization logic, instructions, stimuli, and response formats. MEML compiles this specification into the EML files used by CogGym to render the experiment and construct model prompts. This separates the experimental design from the detailed JSON representation required to execute it. The full specification and workflow are available in the \href{https://coggym.org/doc}{MEML repository}.

We develop an agentic workflow for generating MEML specifications from source papers and original experimental materials. The \texttt{generate-meml} skill guides agents to assemble stimulus, query, instruction, and citation libraries and encode the experimental design as a MEML specification. It then directs agents to compile the specification into EML, validate the output against the EML schemas, verify source-derived design constraints, and check reproducibility across repeated compilations. Source evidence, adaptations, and unresolved design choices are documented for human review; successful compilation alone does not establish fidelity to the original experiment.

Once EMLs are generated, we used an agentic workflow and human inspection to validate EML experiments. We build in two layers of robust checks.

\paragraph{Layer 1: Agentic generation and check.}
We developed an agentic workflow, which compares the MEML specification with the published paper, supplementary materials, original experiment code or survey instruments when available, and released data. The comparison followed a fixed protocol covering ten dimensions: condition assignment, factor structure, trial counts, block structure, ordering and counterbalancing, stimulus content and modality, response type and scale semantics, instructions and cover story, practice/comprehension/feedback modules, and timing or presentation constraints. Agents recorded exact source quotations and locations, confirmations as well as discrepancies, unavailable evidence, and whether a discrepancy could plausibly change participant behavior. Findings and proposed corrections were treated as provisional until human review.

\paragraph{Layer 2: Human experiment audit.}
The human audit had two components. First, the CogGym team conducted an internal audit of the reconstructed experiments. Second, we sought additional review from the original study authors.

Both the researchers conducting the internal audit and the original study authors were asked the same three questions:
\begin{enumerate}
    \item Are the instructions and framing identical to, or close enough to, the original study?
    \item If CogGym presents the stimuli or response options differently from the original study, could this affect participant behavior?
    \item Does the experimental flow or assignment of trials to conditions faithfully match the original study?
\end{enumerate}

We then fix the MEML/EML based on the review and feedback.

\section{Model Prompting and Administration}
\setcounter{subsection}{0}
\subsection{Prompt Construction and Model Harness}
\label{sec:model_eval_details}

The CogGym harness translates each EML trial into a standardized prompt that can be administered to any language model. For each trial, the prompt is constructed as follows:

\paragraph{System prompt.} A fixed preamble instructs the model to behave as an experiment participant:
\begin{lstlisting}[style=eml]
You are a participant in a cognitive science experiment.
Follow the instructions carefully and respond to each
question. Return only the requested JSON object -- do not add text outside the object,
commentary, or caveats. IMPORTANT: You MUST respond in
valid JSON format.
\end{lstlisting}

\paragraph{User message.} The user message concatenates three components:
\begin{enumerate}
    \item \textbf{Experiment instructions}: The condition-specific instructions from \texttt{instruction.jsonl}, with HTML stripped to plain text. Any instruction images are included inline as base64-encoded content.
    \item \textbf{Trial stimuli}: Text descriptions and/or media (images encoded as base64, videos passed natively as base64 or cloud-storage references) from the trial's \texttt{stimuli} array.
    \item \textbf{Queries}: Each query is formatted with its prompt text, response options (for choice types) or scale endpoints (for sliders), and a JSON response template specifying the expected format.
\end{enumerate}

For example, a single-slider query is presented as:
\begin{lstlisting}[style=eml]
Question 1: How much did A cause the outcome?
  Scale: 0=Not at all, 100=Completely
Respond in JSON: {"answer": <number between 0 and 100>,
                  "explanation": "<brief reason>"}
\end{lstlisting}

A multi-choice query:
\begin{lstlisting}[style=eml]
Question 2: What will the agent do next?
Select exactly ONE of the following options:
  A) Go left
  B) Go right
  C) Stay
Respond in JSON: {"answer": "<selected option letter>",
                  "explanation": "<brief reason>"}
\end{lstlisting}

\paragraph{Verbalized-distribution prompt.} To elicit a full response distribution rather than a single answer (\autoref{sec:experiments}), the harness keeps the same system prompt, instructions, and stimuli, but replaces the single-answer response template above with a distribution instruction, following \citet{meister-etal-2025-benchmarking}:
\begin{lstlisting}[style=eml]
=== DISTRIBUTION RESPONSE TASK ===
Do not answer as one participant. Instead, estimate the distribution
of responses that a group of human participants would give.
Return only valid JSON with this exact top-level shape:
{"responses": {"<query_tag>": <distribution_object>}}
Include only the query tags listed below. Probabilities must be numbers
between 0 and 1 and should sum to 1 for each distribution.

Query tags to include:
- cause_rating (single-slider)
  Prompt: How much did A cause the outcome?
  Slider range: 0 to 100
  Return: {"kind": "numeric",
           "support": [{"value": <number>, "prob": <probability>}, ...]}
- action_choice (multi-choice)
  Prompt: What will the agent do next?
  Options: ["Go left", "Go right", "Stay"]
  Return: {"kind": "categorical",
           "probabilities": {"<exact option label>": <probability>, ...}}
\end{lstlisting}
The returned per-option probabilities (categorical) or (value, probability) support (numeric) are then compared directly against the empirical human response distribution.

\section{Evaluation Metrics and Baselines}
\setcounter{subsection}{0}
\subsection{Scoring and Metrics}
\label{sec:metrics}

Model responses are parsed from JSON and scored against human behavioral data using query-type-specific metrics:

\begin{itemize}
    \item \textbf{Single-slider}: The model's numeric response is compared to the human mean. Absolute error (normalized by scale range) and the squared Pearson correlation ($R^2$) across trials measure alignment.
    \item \textbf{Multi-choice}: Across repeated runs, the model's selected options are converted to selection probabilities for each option. Each option--item pair contributes one data point: the model selection probability paired with the human selection proportion. These pairs are pooled within answer type and native scale before computing $R^2$.
    \item \textbf{Multi-select}: The model's binary selection vector is compared to human selection proportions.
    \item \textbf{Multi-slider}: Each sub-item is scored independently as a single-slider.
    \item \textbf{Ranking}: Spearman rank correlation between model and human mean rankings.
\end{itemize}

The primary alignment metric is $\overline{R^2}$: within each experiment, we compute one $R^2$ for each answer type (defined by response regime and native scale), pooling conditions or constructs that share that type and scale; we average equally across answer types and then report the mean across experiments. Human split-half correlations are estimated from 100 random participant splits. For the human $R^2$ references in Tables~\ref{tab:results} and~\ref{tab:subset}, we apply the Spearman--Brown correction $r_{\mathrm{SB}}=2r/(1+r)$ to each experiment's exported mean split-half correlation, square the corrected correlation, and then average across experiments. These corrected references use the 172 recoverable archived exports (80 text, 33 image, 59 video). The human $R^2$ standard errors in Table~\ref{tab:results} are computed across corrected experiment-level values. to ensure reliable estimates, we restrict the split-half computation to response items with at least 10 individual data points. 

\subsection{Random Baseline}
\label{sec:random_baseline}

To place model--human alignment on an interpretable scale, we report a task-naive random baseline, computed over the same items and passed through the identical scoring and normalization used for every model. The baseline ignores the stimulus and draws responses uniformly at random over each item's response space. Because that space is set by the experiment's response format, the randomization adapts to the task type:
\begin{itemize}
    \item \textbf{Continuous (slider) responses:} we draw a value uniformly from the item's admissible range $[\ell, h]$, where $\ell$ and $h$ are the minimum and maximum values the slider permits.
    \item \textbf{Categorical (forced-choice) responses:} we select one of the $k$ options uniformly at random, yielding a one-hot vector over the options.
\end{itemize}
For the $R^2$ baseline, we repeat the draw $1{,}000$ times with a fixed random seed and compute the same per-experiment $R^2$ against the human means as for the models. For the divergence baseline, we compare the full task-naive uniform response distribution directly with the empirical human response distribution, using normalized Wasserstein-1 distance over the admissible response range for continuous items and normalized Jensen--Shannon distance over the available options for categorical items. This distribution-to-distribution comparison matches the metric used for model responses; it is not the average divergence of individual random point responses. We average item scores within each answer type and native scale, then equally across answer types within an experiment and across experiments within a modality. The resulting values populate the ``Random baseline'' row of \autoref{tab:results} and the dotted reference lines in Figures~\ref{fig:gpqa_marquee} and~\ref{fig:scaling_r2}.

\section{Model--Human Alignment Results}
\setcounter{subsection}{0}
\subsection{Full Model--Human Alignment Results}
\label{sec:full_results}

\autoref{tab:results} reports per-model human-alignment scores, measured by $R^2$ (central-tendency alignment) and normalized distribution divergence, broken down by modality.

\begin{table}[H]
\centering
\scriptsize
\setlength{\tabcolsep}{2pt}
\renewcommand{\arraystretch}{0.6}
\begin{tabular*}{\linewidth}{@{\extracolsep{\fill}}p{1.35in} *{6}{>{\centering\arraybackslash}p{0.46in}}}
\toprule
& \multicolumn{3}{c}{\textbf{$R^2$}~{\small$\uparrow$}} & \multicolumn{3}{c}{\textbf{Norm.\ Div.}~{\small$\downarrow$}} \\
\cmidrule(lr){2-4}\cmidrule(lr){5-7}
\textbf{Model} & Text{\tiny\,(91)} & Image{\tiny\,(82)} & Video{\tiny\,(85)} & Text{\tiny\,(86)} & Image{\tiny\,(80)} & Video{\tiny\,(67)} \\
\midrule  Human reference$^{\dagger}$ & 0.93{\tiny$\pm$.01} & 0.95{\tiny$\pm$.01} & 0.92{\tiny$\pm$.01} & 0.05{\tiny$\pm$.01} & 0.07{\tiny$\pm$.01} & 0.07{\tiny$\pm$.01} \\
Random baseline & 0.07{\tiny$\pm$.02} & 0.05{\tiny$\pm$.01} & 0.03{\tiny$\pm$.01} & 0.28{\tiny$\pm$.01} & 0.26{\tiny$\pm$.01} & 0.29{\tiny$\pm$.01} \\
\midrule
\multicolumn{7}{@{}l}{Claude (Anthropic)} \\
\hspace{0.8em}Claude Opus 4.7 & 0.55{\tiny$\pm$.03} & 0.45{\tiny$\pm$.04} & -- & 0.10{\tiny$\pm$.01} & 0.17{\tiny$\pm$.01} & -- \\
\hspace{0.8em}Claude Sonnet 4.5 & 0.52{\tiny$\pm$.03} & 0.38{\tiny$\pm$.03} & -- & 0.11{\tiny$\pm$.01} & 0.20{\tiny$\pm$.01} & -- \\
\hspace{0.8em}Claude Haiku 4.5 & 0.46{\tiny$\pm$.03} & 0.22{\tiny$\pm$.03} & -- & 0.13{\tiny$\pm$.01} & 0.20{\tiny$\pm$.01} & -- \\
\midrule
\multicolumn{7}{@{}l}{GPT (OpenAI)} \\
\hspace{0.8em}GPT-5.2 & 0.51{\tiny$\pm$.03} & 0.47{\tiny$\pm$.03} & -- & 0.11{\tiny$\pm$.01} & 0.21{\tiny$\pm$.02} & -- \\
\hspace{0.8em}GPT-5 mini & 0.47{\tiny$\pm$.03} & 0.47{\tiny$\pm$.03} & -- & 0.13{\tiny$\pm$.01} & 0.20{\tiny$\pm$.02} & -- \\
\hspace{0.8em}GPT-4.1 & 0.49{\tiny$\pm$.03} & 0.33{\tiny$\pm$.04} & -- & 0.12{\tiny$\pm$.01} & 0.21{\tiny$\pm$.02} & -- \\
\hspace{0.8em}GPT-4o & 0.52{\tiny$\pm$.03} & 0.30{\tiny$\pm$.03} & -- & 0.13{\tiny$\pm$.01} & 0.21{\tiny$\pm$.02} & -- \\
\hspace{0.8em}GPT-4o mini & 0.43{\tiny$\pm$.03} & 0.21{\tiny$\pm$.03} & -- & 0.15{\tiny$\pm$.01} & 0.24{\tiny$\pm$.02} & -- \\
\hspace{0.8em}GPT-OSS 120B & 0.49{\tiny$\pm$.03} & -- & -- & 0.13{\tiny$\pm$.01} & -- & -- \\
\hspace{0.8em}GPT-OSS 20B & 0.46{\tiny$\pm$.03} & -- & -- & 0.14{\tiny$\pm$.01} & -- & -- \\
\midrule
\multicolumn{7}{@{}l}{Gemini (Google)} \\
\hspace{0.8em}Gemini 3.1 Pro & 0.59{\tiny$\pm$.03} & 0.58{\tiny$\pm$.03} & 0.43{\tiny$\pm$.03} & 0.11{\tiny$\pm$.01} & 0.17{\tiny$\pm$.01} & 0.25{\tiny$\pm$.02} \\
\hspace{0.8em}Gemini 3 Flash & 0.52{\tiny$\pm$.03} & 0.53{\tiny$\pm$.03} & 0.28{\tiny$\pm$.02} & 0.11{\tiny$\pm$.01} & 0.17{\tiny$\pm$.01} & 0.28{\tiny$\pm$.02} \\
\hspace{0.8em}Gemini 2.5 Pro & 0.51{\tiny$\pm$.03} & 0.52{\tiny$\pm$.03} & 0.25{\tiny$\pm$.02} & 0.12{\tiny$\pm$.01} & 0.18{\tiny$\pm$.01} & 0.32{\tiny$\pm$.02} \\
\hspace{0.8em}Gemini 2.5 Flash & 0.51{\tiny$\pm$.03} & 0.46{\tiny$\pm$.03} & 0.19{\tiny$\pm$.02} & 0.13{\tiny$\pm$.01} & 0.19{\tiny$\pm$.02} & 0.35{\tiny$\pm$.02} \\
\midrule
\multicolumn{7}{@{}l}{Qwen (Alibaba)} \\
\hspace{0.8em}Qwen3.5 Flash 35B & 0.49{\tiny$\pm$.03} & 0.39{\tiny$\pm$.03} & 0.13{\tiny$\pm$.03} & 0.15{\tiny$\pm$.01} & 0.21{\tiny$\pm$.02} & 0.27{\tiny$\pm$.02} \\
\hspace{0.8em}Qwen3.5 9B & 0.35{\tiny$\pm$.03} & 0.19{\tiny$\pm$.03} & 0.10{\tiny$\pm$.02} & 0.18{\tiny$\pm$.01} & 0.23{\tiny$\pm$.02} & 0.30{\tiny$\pm$.02} \\
\hspace{0.8em}Qwen3.5 4B & 0.30{\tiny$\pm$.03} & 0.19{\tiny$\pm$.03} & 0.09{\tiny$\pm$.02} & 0.18{\tiny$\pm$.01} & 0.25{\tiny$\pm$.02} & 0.31{\tiny$\pm$.02} \\
\hspace{0.8em}Qwen3.5 2B & 0.12{\tiny$\pm$.01} & 0.08{\tiny$\pm$.02} & 0.05{\tiny$\pm$.01} & 0.23{\tiny$\pm$.02} & 0.25{\tiny$\pm$.02} & 0.30{\tiny$\pm$.02} \\
\hspace{0.8em}Qwen3-VL 32B & 0.49{\tiny$\pm$.03} & 0.26{\tiny$\pm$.03} & 0.12{\tiny$\pm$.02} & 0.14{\tiny$\pm$.01} & 0.22{\tiny$\pm$.02} & 0.30{\tiny$\pm$.02} \\
\hspace{0.8em}Qwen3-VL 8B & 0.37{\tiny$\pm$.03} & 0.22{\tiny$\pm$.03} & 0.11{\tiny$\pm$.02} & 0.16{\tiny$\pm$.01} & 0.23{\tiny$\pm$.02} & 0.30{\tiny$\pm$.02} \\
\hspace{0.8em}Qwen3-VL 4B & 0.33{\tiny$\pm$.03} & 0.15{\tiny$\pm$.03} & 0.10{\tiny$\pm$.02} & 0.17{\tiny$\pm$.01} & 0.25{\tiny$\pm$.02} & 0.32{\tiny$\pm$.02} \\
\hspace{0.8em}Qwen2.5-VL 32B & 0.47{\tiny$\pm$.03} & 0.25{\tiny$\pm$.03} & 0.08{\tiny$\pm$.01} & 0.14{\tiny$\pm$.01} & 0.22{\tiny$\pm$.02} & 0.29{\tiny$\pm$.02} \\
\hspace{0.8em}Qwen2.5-VL 7B & 0.32{\tiny$\pm$.03} & 0.12{\tiny$\pm$.02} & 0.07{\tiny$\pm$.01} & 0.17{\tiny$\pm$.01} & 0.24{\tiny$\pm$.02} & 0.28{\tiny$\pm$.02} \\
\hspace{0.8em}Qwen2.5-VL 3B & 0.26{\tiny$\pm$.03} & 0.10{\tiny$\pm$.02} & 0.05{\tiny$\pm$.01} & 0.25{\tiny$\pm$.02} & 0.24{\tiny$\pm$.01} & 0.29{\tiny$\pm$.02} \\
\hspace{0.8em}Qwen3.5 0.8B & 0.08{\tiny$\pm$.01} & 0.06{\tiny$\pm$.01} & 0.05{\tiny$\pm$.01} & 0.31{\tiny$\pm$.02} & 0.43{\tiny$\pm$.07} & 0.32{\tiny$\pm$.02} \\
\hspace{0.8em}Qwen3 1.7B & 0.23{\tiny$\pm$.03} & -- & -- & 0.23{\tiny$\pm$.06} & -- & -- \\
\hspace{0.8em}Qwen2.5 1.5B & 0.15{\tiny$\pm$.02} & -- & -- & 0.26{\tiny$\pm$.02} & -- & -- \\
\hspace{0.8em}Qwen2.5 0.5B & 0.11{\tiny$\pm$.02} & -- & -- & 0.33{\tiny$\pm$.04} & -- & -- \\
\midrule
\multicolumn{7}{@{}l}{Llama (Meta)} \\
\hspace{0.8em}Llama 4 Maverick & 0.40{\tiny$\pm$.03} & 0.23{\tiny$\pm$.03} & -- & 0.14{\tiny$\pm$.01} & 0.24{\tiny$\pm$.02} & -- \\
\hspace{0.8em}Llama 4 Scout & 0.39{\tiny$\pm$.03} & 0.22{\tiny$\pm$.03} & -- & 0.16{\tiny$\pm$.01} & 0.23{\tiny$\pm$.02} & -- \\
\hspace{0.8em}Llama 3.3 70B & 0.46{\tiny$\pm$.03} & -- & -- & 0.14{\tiny$\pm$.01} & -- & -- \\
\hspace{0.8em}Llama 3.2 11B & 0.32{\tiny$\pm$.03} & 0.22{\tiny$\pm$.07} & -- & 0.23{\tiny$\pm$.02} & 0.27{\tiny$\pm$.02} & -- \\
\hspace{0.8em}Llama 3.2 3B & 0.25{\tiny$\pm$.03} & -- & -- & 0.20{\tiny$\pm$.03} & -- & -- \\
\hspace{0.8em}Llama 3.2 1B & 0.10{\tiny$\pm$.02} & -- & -- & 0.24{\tiny$\pm$.03} & -- & -- \\
\hspace{0.8em}Llama 3.1 70B & 0.48{\tiny$\pm$.03} & -- & -- & 0.14{\tiny$\pm$.01} & -- & -- \\
\hspace{0.8em}Llama 3.1 8B & 0.35{\tiny$\pm$.03} & -- & -- & 0.22{\tiny$\pm$.01} & -- & -- \\
\midrule
\multicolumn{7}{@{}l}{Gemma (Google)} \\
\hspace{0.8em}Gemma 4 26B & 0.45{\tiny$\pm$.03} & 0.27{\tiny$\pm$.03} & -- & 0.13{\tiny$\pm$.01} & 0.24{\tiny$\pm$.02} & -- \\
\hspace{0.8em}Gemma 4 E2B & 0.28{\tiny$\pm$.03} & -- & -- & 0.24{\tiny$\pm$.02} & -- & -- \\
\hspace{0.8em}Gemma 3 27B & 0.43{\tiny$\pm$.03} & 0.22{\tiny$\pm$.03} & -- & 0.13{\tiny$\pm$.01} & 0.21{\tiny$\pm$.02} & -- \\
\hspace{0.8em}Gemma 3 12B & 0.37{\tiny$\pm$.03} & -- & -- & 0.17{\tiny$\pm$.01} & -- & -- \\
\hspace{0.8em}Gemma 3 4B & 0.26{\tiny$\pm$.03} & -- & -- & 0.19{\tiny$\pm$.01} & -- & -- \\
\midrule
\multicolumn{7}{@{}l}{Other} \\
\hspace{0.8em}Grok 4.3 & 0.52{\tiny$\pm$.03} & 0.30{\tiny$\pm$.04} & -- & 0.13{\tiny$\pm$.01} & 0.20{\tiny$\pm$.02} & -- \\
\hspace{0.8em}GLM 5.2 & 0.53{\tiny$\pm$.03} & -- & -- & 0.10{\tiny$\pm$.01} & -- & -- \\
\hspace{0.8em}Kimi K2.5 & 0.48{\tiny$\pm$.03} & 0.34{\tiny$\pm$.03} & -- & 0.10{\tiny$\pm$.01} & 0.15{\tiny$\pm$.01} & -- \\
\hspace{0.8em}Mistral Small 24B & 0.44{\tiny$\pm$.03} & -- & -- & 0.15{\tiny$\pm$.01} & -- & -- \\
\hspace{0.8em}Mistral 7B & 0.30{\tiny$\pm$.03} & -- & -- & 0.19{\tiny$\pm$.01} & -- & -- \\
\hspace{0.8em}DeepSeek V4 Pro & 0.50{\tiny$\pm$.03} & -- & -- & 0.13{\tiny$\pm$.01} & -- & -- \\
\hspace{0.8em}DeepSeek V4 Flash & 0.45{\tiny$\pm$.03} & -- & -- & 0.13{\tiny$\pm$.01} & -- & -- \\
\hspace{0.8em}DeepSeek V3.2 & 0.40{\tiny$\pm$.03} & -- & -- & 0.13{\tiny$\pm$.01} & -- & -- \\
\hspace{0.8em}Centaur~\citep{centaur2024} & 0.42{\tiny$\pm$.03} & -- & -- & 0.19{\tiny$\pm$.01} & -- & -- \\
\bottomrule
\end{tabular*}
\caption{\textbf{Model--human alignment measured by $R^2$ and normalized distribution divergence.} All models use default configuration with thinking mode enabled if available. Values following $\pm$ show standard errors. $^{\dagger}$Human $R^2$ values are Spearman--Brown-corrected references from the 172 recoverable archived exports (80 text, 33 image, 59 video); their standard errors are computed across corrected experiment-level $R^2$ values. Parenthetical counts in the header give the number of benchmark experiments eligible for each metric before model-specific missingness. Centaur is the Llama 3.1 70B base model fine-tuned on human behavioral data; all other models are instruction-tuned.}
\label{tab:results}
\end{table}

\subsection{Model--Human \texorpdfstring{$R^2$}{R2} on Split-Half Reliable Subset}
\label{sec:splithalf_subset}

\autoref{tab:subset} reports the top 10 models by text $R^2$ on the 179 experiments (text: 80, image: 37, video: 62) where human split-half reliability can be reliably estimated (average $\geq$10 individual responses per item). The Spearman--Brown-corrected human reference is high (text: $R^2=0.932$, image: $0.945$, video: $0.918$), indicating strong human agreement. We corrected each experiment's exported mean split-half correlation as $r_{\mathrm{SB}}=2r/(1+r)$ before squaring and averaging across experiments. This human reference uses the 172 recoverable archived exports (80 text, 33 image, 59 video), as in \autoref{fig:gpqa_marquee}; the 179-experiment model subset is unchanged. Grok 4.3 leads on text (0.60), while Gemini 3.1 Pro leads on image (0.53) and video (0.23). The gap between the best model and human reliability remains substantial across all modalities.

\begin{table}[h]
\centering
\small
\begin{tabular*}{\linewidth}{@{\extracolsep{\fill}}lccc}
\toprule
& \textbf{Text} & \textbf{Image} & \textbf{Video} \\
\textbf{Model} & \textbf{$R^2$ $\uparrow$} & \textbf{$R^2$ $\uparrow$} & \textbf{$R^2$ $\uparrow$} \\
\midrule
Grok 4.3 & 0.60 & 0.19 & --- \\
Gemini 3.1 Pro & 0.59 & 0.53 & 0.23 \\
Claude Opus 4.7 & 0.58 & 0.42 & --- \\
Claude Sonnet 4.5 & 0.58 & 0.24 & --- \\
Qwen3.5 Flash 35B & 0.57 & 0.38 & --- \\
GPT-5.2 & 0.55 & 0.46 & --- \\
GPT-5 mini & 0.54 & 0.38 & --- \\
Kimi K2.5 & 0.52 & 0.29 & --- \\
GPT-4o & 0.51 & 0.28 & --- \\
Gemma 4 26B & 0.51 & 0.26 & --- \\
\midrule  Human (SB-corrected) & 0.93 & 0.95 & 0.92 \\
\bottomrule
\end{tabular*}
\caption{\textbf{Top 10 model--human $R^2$ on the split-half reliable subset} (179 experiments), ranked by text $R^2$. The human reference is Spearman--Brown-corrected using the 172 recoverable archived exports (80 text, 33 image, 59 video).}
\label{tab:subset}
\end{table}

\section{Model Sampling and Response Variability}
\setcounter{subsection}{0}
\subsection{Run-Count Sensitivity Analysis}
\label{sec:run_count_sensitivity}

A possible concern is that too few model samples were collected per item. To test this directly, we recomputed Gemma 4 26B text $R^2$ using increasing numbers of completed runs while holding prompts, temperature, and trial selection fixed. Additional runs stabilize the estimate but do not change the conclusion: mean $R^2$ remains effectively flat, with overlapping 95\% confidence intervals across all run counts (\autoref{tab:run_count_sensitivity}).

\begin{table}[h]
\centering
\small
\begin{tabular*}{\linewidth}{@{\extracolsep{\fill}}rcc}
\toprule
\textbf{Runs} & \textbf{Mean $R^2$} & \textbf{95\% CI} \\
\midrule
5  & 0.49 & [0.43, 0.55] \\
10 & 0.48 & [0.42, 0.54] \\
15 & 0.48 & [0.42, 0.53] \\
20 & 0.49 & [0.43, 0.55] \\
25 & 0.48 & [0.42, 0.54] \\
30 & 0.48 & [0.42, 0.54] \\
35 & 0.48 & [0.42, 0.54] \\
40 & 0.48 & [0.42, 0.54] \\
45 & 0.48 & [0.42, 0.54] \\
50 & 0.48 & [0.42, 0.54] \\
\bottomrule
\end{tabular*}
\caption{\textbf{Effect of number of model runs on Gemma 4 26B text alignment.} Increasing the number of runs from 5 to 50 does not materially improve $R^2$; the 95\% confidence intervals (1000-sample bootstrap over experiments) overlap across all run counts.}
\label{tab:run_count_sensitivity}
\end{table}

\subsection{Sampled Responses Under-express Human Variability}
\label{sec:peaked}

\begin{figure}[h]
    \centering
    \includegraphics[width=\textwidth]{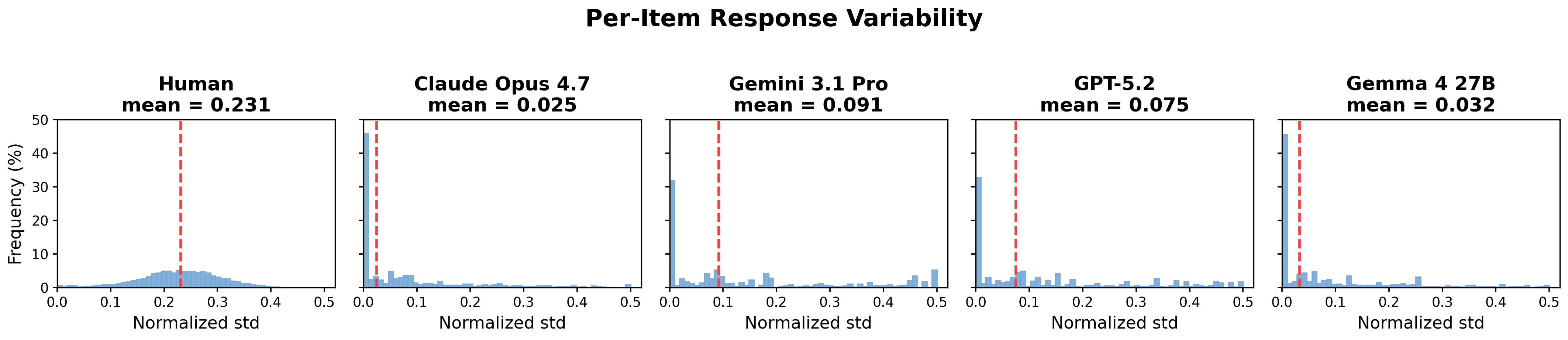}
    \caption{\textbf{Response diversity: humans vs.\ repeated model runs.} For each item, response variability is computed on the normalized $[0,1]$ response scale, and the distributions are compared across humans and repeated model runs. Values near $0$ indicate nearly deterministic responses; broader distributions indicate greater item-level variability. Sampled model responses are markedly less diverse than human responses.}
    \label{fig:diversity}
\end{figure}

A central motivation for eliciting model-predicted human response distributions through verbalized prompting~\citep{meister-etal-2025-benchmarking} rather than repeated sampling is that, when sampled at temperature $1.0$, model responses are sharply peaked relative to humans: repeated runs concentrate on a single high-probability response and recover only a fraction of the item-level variability present across human participants. \autoref{fig:diversity} visualizes this---for each item we compute response variability on the normalized response scale and compare the resulting distributions across humans and repeated model runs (values near $0$ indicate nearly deterministic responses). Sampled diversity varies markedly across models: GPT-5.2 is the most diverse non-thinking model ($72\%$ of human variability), followed by Gemini 3.1 Pro ($38\%$), Gemma 4 ($14\%$), and Claude Opus ($10\%$)---but all fall well short of human spread. Repeated sampling thus captures only a fraction of the item-level variability present across human participants, even for models that track the human mean reasonably well.

\begin{figure}[h]
    \centering
    \includegraphics[width=\textwidth]{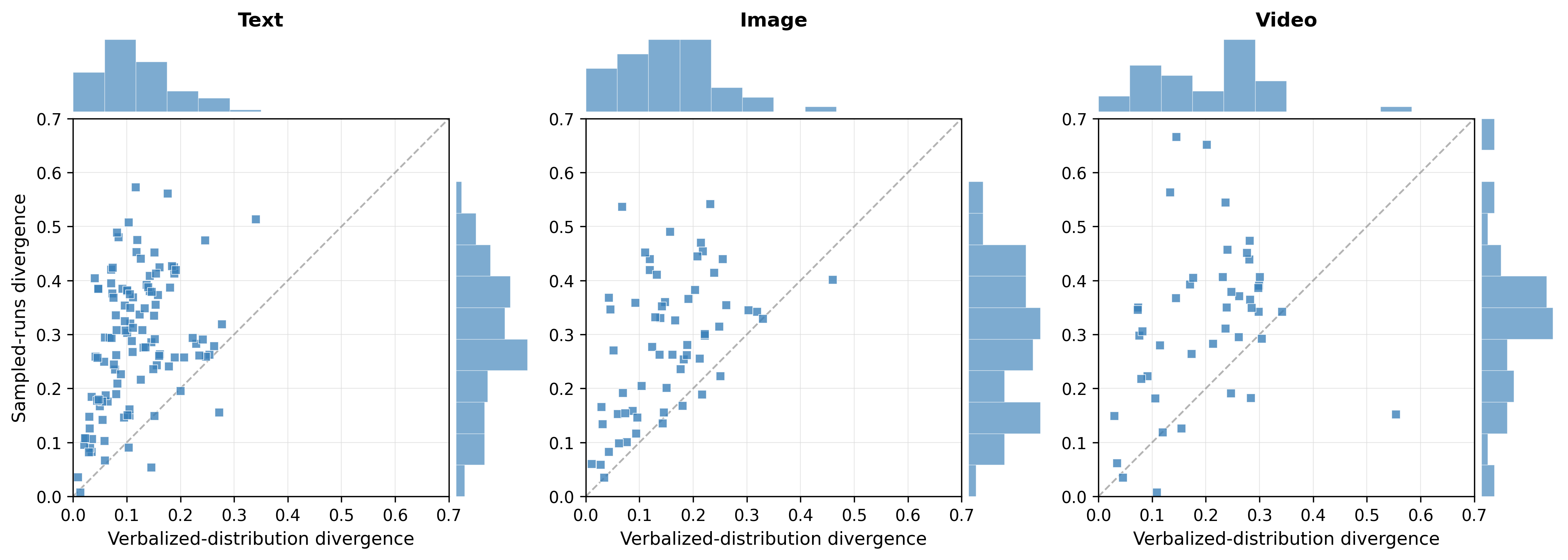}
    \caption{\textbf{Verbalized-distribution vs.\ sampled-runs divergence (Gemini 3.1 Pro).} Each point is one experiment; $x$ is the normalized divergence from verbalized elicitation and $y$ from $10$ sampled runs (lower is better). The vast majority of experiments fall above the parity diagonal, i.e., repeated sampling---being sharply peaked---diverges from the human response distribution more than the verbalized estimate does.}
    \label{fig:verb_vs_sampled}
\end{figure}

This peakedness directly inflates distributional divergence. \autoref{fig:verb_vs_sampled} compares, per experiment for Gemini 3.1 Pro, the normalized divergence obtained from repeated sampling ($10$ runs) against that obtained from verbalized elicitation. Sampling yields substantially larger divergence in almost every experiment---the median rises from $0.11$ to $0.29$ on text, $0.14$ to $0.29$ on image, and $0.23$ to $0.34$ on video, with $91\%$/$88\%$/$78\%$ of experiments above the parity diagonal. Repeated sampling and verbalized elicitation probe distinct quantities: the former estimates variability in the model's own outputs across runs, whereas the latter elicits its explicit prediction of the human response distribution. Because our distributional-alignment analysis asks how well models predict human response variability, we use verbalized elicitation for all reported divergence computations.

\section{Human Replications}
\setcounter{subsection}{0}
\label{sec:replication_bound}
A natural concern with any large-scale benchmark derived from published experiments is whether poor model--human alignment reflects genuine cognitive divergence or artifacts of data standardization.

Here, $R^2$ denotes squared Pearson correlation between response means, and the overall summaries are unweighted arithmetic means of the 15 experiment-level values. Mean replication--original agreement is $R^2 = 0.84$, compared with $R^2 = 0.62$ for Gemini~3.1 Pro versus the original humans. 

\begin{table}[h]
\centering
\small
\begin{tabular*}{\linewidth}{@{\extracolsep{\fill}}llccc}
\toprule
\textbf{Experiment} & \textbf{Modality} & $\boldsymbol{N}$ & \textbf{Repl.}~$\boldsymbol{R^2}$ & \textbf{G3.1\,Pro}~$\boldsymbol{R^2}$ \\
\midrule
\href{https://coggym.org/studies/levine2020logic-exp1}{Levine (2020), Exp.~1}          & Text  & 35 & 0.86 & 0.48 \\

\href{https://coggym.org/studies/tsvilodub2025nonliteral-exp1}{Tsvilodub (2025), Exp.~1}  & Text  & 30 & 0.79 & 0.39 \\

\href{https://coggym.org/studies/ying2023neuro-exp1}{Ying (2023), Exp.~1}            & Text  & 30 & 0.82 & 0.70 \\
\href{https://coggym.org/studies/yoon2020polite-exp1}{Yoon (2020), Exp.~1}             & Text  & 30 & 0.97 & 0.95 \\
\href{https://coggym.org/studies/hu2023fine-exp1}{Hu (2023), Exp.~1}                 & Text  & 30 & 0.99 & 0.93 \\
\midrule
\href{https://coggym.org/studies/aboody2025inferring-exp2}{Aboody (2025), Exp.~2}      & Image & 30 & 0.89 & 0.84 \\
\href{https://coggym.org/studies/jara-ettinger2020naive-exp6}{Jara-Ettinger (2020), Exp.~6}   & Image & 30 & 0.80 & 0.81 \\
\href{https://coggym.org/studies/lopez-brau2023people-exp1}{Lopez-Brau (2023), Exp.~1}     & Image & 50 & 0.75 & 0.78 \\
\href{https://coggym.org/studies/jaraettinger2021quantitative-exp2}{Jara-Ettinger (2021), Exp.~2} & Image & 30 & 0.77 & 0.35 \\
\href{https://coggym.org/studies/chandra2024cooperative-exp1}{Chandra (2024), Exp.~1} & Image & 67 & 0.73 & 0.62 \\
\midrule
\href{https://coggym.org/studies/fu2025hierarchical-exp1}{Fu (2025), Exp.~1}       & Video & 30 & 0.92 & 0.66 \\
\href{https://coggym.org/studies/bass2022partial-exp1}{Bass (2022), Exp.~1}          & Video & 39 & 0.92 & 0.21 \\
\href{https://coggym.org/studies/sosa2021moral-exp1}{Sosa (2021), Exp.~1}            & Video & 30 & 0.82 & 0.15 \\
\href{https://doi.org/10.31234/osf.io/uwdbr}{Wu (2023), Exp.~1b}  & Video & 30 & 0.81 & 0.90 \\
\href{https://coggym.org/studies/royka2022people-exp1}{Royka (2022), Exp.~1}          & Video & 30 & 0.75 & 0.56 \\
\midrule
\textbf{Mean / total} & & \textbf{521} & \textbf{0.84} & \textbf{0.62}
 \\

\bottomrule
\end{tabular*}
\caption{\textbf{Experiment replication and comparisons.} $N$ is the number of participants in each included replication batch; the total sums study participations. \textbf{Repl.}~$R^2$ compares replication means against original human means; \textbf{G3.1\,Pro}~$R^2$ compares Gemini~3.1 Pro against the original humans.}
\label{tab:replication}
\end{table}


\section{Cross-Model Comparison}
\setcounter{subsection}{0}
\label{sec:cross_model_correlation}

\begin{figure}[h]
    \centering
    \includegraphics[width=\textwidth]{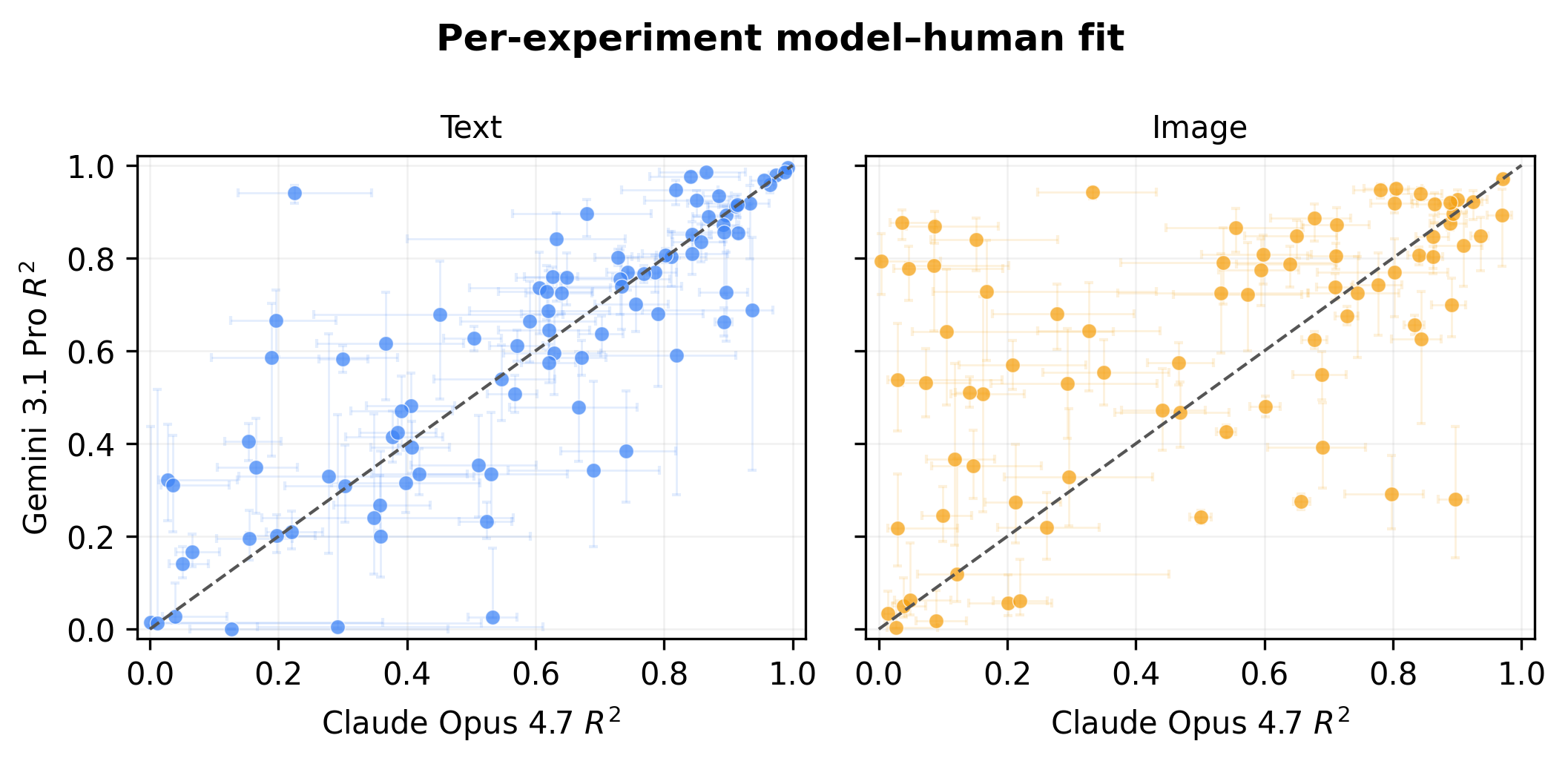}
    \caption{\textbf{Per-experiment $R^2$: Claude Opus 4.7 vs.\ Gemini 3.1 Pro.} Each point is one experiment. Overall the plots show strong correlation between the two models. Error bars on both axes are 95\% confidence intervals (1000-sample bootstrap percentile, resampling items); experiments whose $R^2$ is too poorly estimated to compare (95\% CI width $>0.4$ on either axis, typically those with very few items) are omitted.}
    \label{fig:opus_vs_g31pro}
\end{figure}

Model behavior is strongly correlated across experiments: Claude Opus 4.7 and Gemini 3.1 Pro tend to succeed and fail on the same experiments (\autoref{fig:opus_vs_g31pro}), though notable divergences remain where one model is well aligned and the other near chance.

This cross-model consistency carries two implications. First, per-experiment difficulty is largely a property of the task rather than the model: an experiment that one frontier model struggles on is likely to be difficult for the other as well, so the alignment gaps we report reflect shared inductive limitations---systematically over-confident, under-dispersed judgments relative to graded human responses---rather than idiosyncratic errors of a single system. Second, because frontier models rise and fall together across CogGym, their relative ordering is more stable than the absolute alignment level, which remains far below the human ceiling for all of them; swapping model families is therefore unlikely, on its own, to close the gap. The off-diagonal experiments---where the two models disagree---are the most informative for diagnosis, marking paradigms where the models bring different priors to bear, and are natural candidates for targeted follow-up.

\section{Contamination Analysis}
\label{sec:study_recognition}

\paragraph{Procedure}
We conducted a one-run, memory-only study-recognition probe on 175 experiments (92 text and 83 static-image experiments, spanning 59 papers). Each independent request presented anonymized instructions, stimulus examples, and questions. Models were asked to recall the paper title, authors, publication year, and any formal model, and to report separate confidence estimates for paper identification and model recall. No web search, retrieval tools were provided. The request contained up to 40 trial examples. For image experiments, we additionally attached original images for up to three trial examples without any human response data. 

\paragraph{Edit-distance title similarity}
For the continuous-similarity analysis reported in Results, we compute similarity between the original title and paper title provided by AI as $1-d(a,b)/\max(|a|,|b|)$, where $d$ is character-level Levenshtein distance with unit costs for insertion, deletion, and substitution, and lengths are measured after normalization. This score ranges from zero to one, with one indicating identical normalized titles.

\paragraph{Model prompt}
The following system prompt was shared across models. The user message supplied only the anonymized, experiment-specific materials described above; it did not include ground-truth bibliographic metadata or human responses. The prompt's claim that identifying details were removed describes the intended anonymization, subject to the limitations noted above.
\begin{lstlisting}[basicstyle=\ttfamily\scriptsize,breaklines=true,columns=fullflexible,keepspaces=true,showstringspaces=false]
# Knowledge check -- identify this study from memory

You are shown the (anonymized) materials of a single published cognitive-science experiment: its instructions, stimuli, and questions, as they were presented to human participants. All identifying details -- the paper's title, authors, institution, and any human data -- have been removed.

Your job is to recognize the study **from your own memory alone**. You have no internet access and no search tools; do not attempt to look anything up. If you are unsure, give your best guess and a low confidence rather than refusing.

Answer these questions:

1. **Which published study is this?** Give the paper's title, its authors, and the year of publication -- your best recollection.
2. **Did that paper introduce or test a formal / computational model?** If so, name it and briefly describe it from memory: the model family, and its key parameters or mechanism. If the paper introduced no computational model, say so (`has_model: false`).
3. **How confident are you?** Give a confidence in [0, 1] for the paper identification and a separate confidence for the model recall.

Reason from the specific structure of the task -- the cover story, the manipulated variables, the response format, the number and design of trials -- to the paper you remember matching it. Base every answer only on what you already know; never invent a citation you are not actually recalling (an honest "unknown" with low confidence is more useful than a fabricated title).

Reply with ONLY one JSON object, no prose before or after it, in exactly this shape:

{
  "paper_title": "<the paper's title, best guess from memory, or \"\" if unknown>",
  "authors": "<author surnames you recall, or \"\">",
  "year": "<publication year (YYYY), or \"unknown\">",
  "paper_confidence": 0.0,
  "has_model": false,
  "model_name": "<name of the computational model the paper introduced/tested, or null>",
  "model_description": "<from memory: model family, key parameters/mechanism, or \"\">",
  "model_confidence": 0.0
}

`paper_title` and `has_model` are required; the rest are optional. Confidences are in [0, 1].
\end{lstlisting}

\clearpage
\section{Severely Misaligned Cases}
\label{sec:failure_cases}

We selected trials on which the strongest models disagree with a clear human consensus, drawing from experiments that span different topics and both the text and image modalities. Candidates were identified by ranking every scored item by the scale-normalized gap between the model mean and the human mean, weighted by human agreement, for Gemini~3.1~Pro, Claude~Opus~4.7, and GPT-5.2; experiments with known stimulus-delivery or scoring artifacts were excluded. Each figure shows the stimulus and question as presented, followed by the distribution of the original participants' responses and, in separate panels, the distribution of each model's sampled responses on the same trial ($10$ runs per model). The examples are illustrative rather than representative.

\begin{figure}[h]
    \centering
    \includegraphics[width=\textwidth]{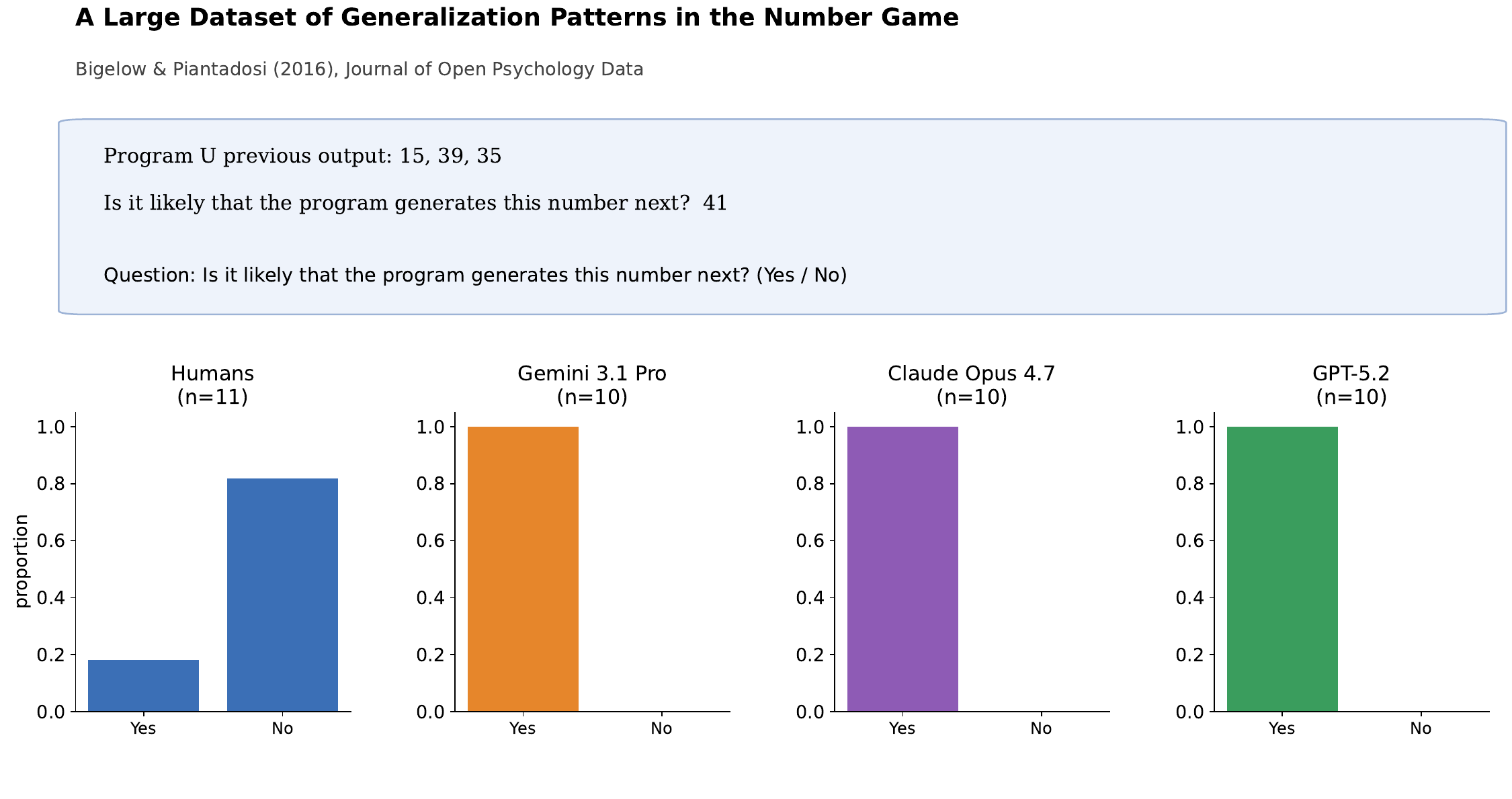}
    \caption{\textbf{Concept learning (text): the number game} \citep{bigelow2016large}. Participants see a few outputs of an unknown program and judge whether a new number is likely to be generated next. Given \emph{15, 39, 35}, only $18\%$ of participants ($n=11$) say that \emph{41} is likely, but every run of every model says yes, with the same justification: \emph{\textquotedblleft{}All previous numbers are odd integers, and 41 is also an odd integer.\textquotedblright{}} Human generalization in this task hedges between rule-based hypotheses (odd numbers) and similarity-based ones (numbers in the teens and thirties); the models commit to the first compact rule that fits.}
    \label{fig:failure_bigelow}
\end{figure}

\begin{figure}[h]
    \centering
    \includegraphics[width=\textwidth]{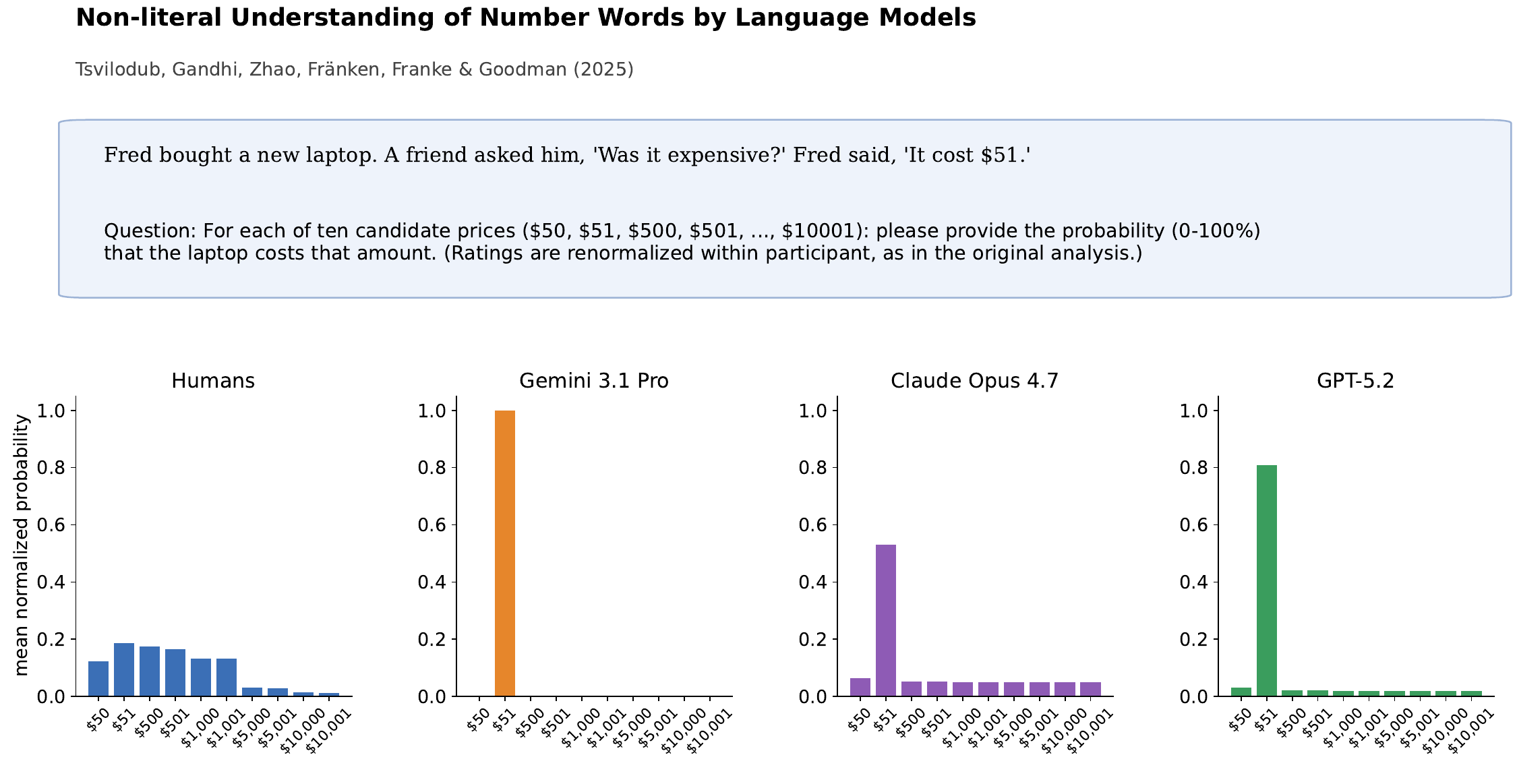}
    \caption{\textbf{Pragmatics of number words (text)} \citep{tsvilodub2025nonliteral}. After reading \emph{\textquotedblleft{}It cost \$51,\textquotedblright{}} participants rate the probability of ten candidate prices; ratings are renormalized within participant as in the original analysis. Humans spread their probability across \$50, \$51, \$500, \$501, \$1{,}000, and \$1{,}001, placing only about $19\%$ on the literal \$51 ($n=66$). Gemini places $100\%$ on the literal value in every run and GPT-5.2 about $80\%$ on average (\emph{\textquotedblleft{}Fred explicitly stated the price was \$51, indicating a precise and literal value\textquotedblright{}}). Opus splits: in five of ten runs it places $90$--$95\%$ on \$51, and in the other five it returns the same value for every price, which renormalizes to a flat distribution. None of the models reproduces the human pattern of treating the number as an imprecise report.}
    \label{fig:failure_tsvilodub}
\end{figure}

\begin{figure}[h]
    \centering
    \includegraphics[width=\textwidth]{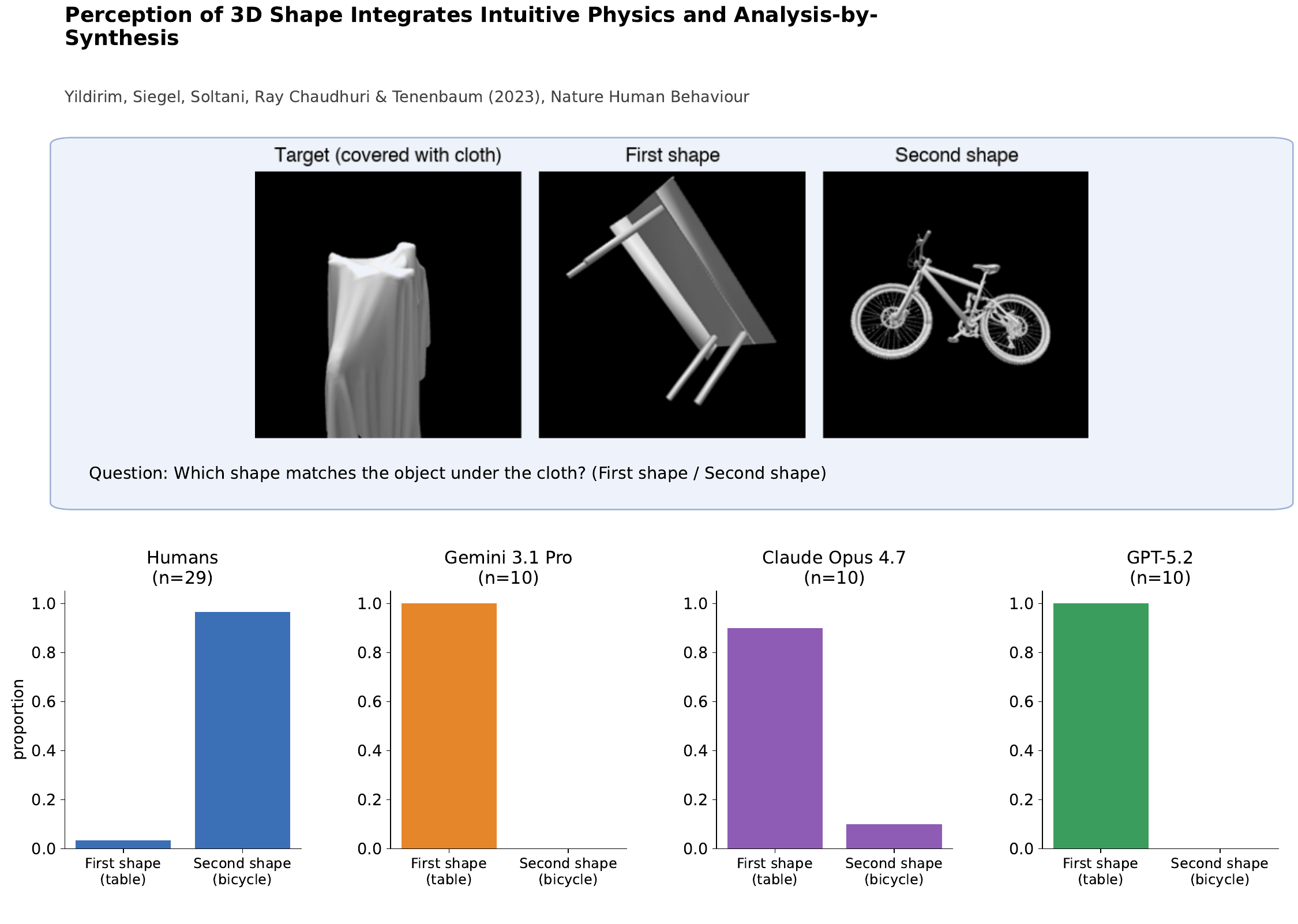}
    \caption{\textbf{3D shape perception (image): reading the silhouette instead of the physics} \citep{yildirim2024perception}. Participants match a cloth-covered target to one of two uncovered objects, a task people solve by reasoning about how cloth settles on a shape. $97\%$ of participants ($n=29$) identify the bicycle; Gemini and GPT-5.2 choose the table in every run and Opus in nine of ten, describing the draped outline as if it were the object: \emph{\textquotedblleft{}The draped cloth outlines a flat rectangular surface with four protruding legs, which perfectly matches the structure of the table\textquotedblright{}} (Gemini).}
    \label{fig:failure_yildirim}
\end{figure}

\begin{figure}[h]
    \centering
    \includegraphics[width=\textwidth]{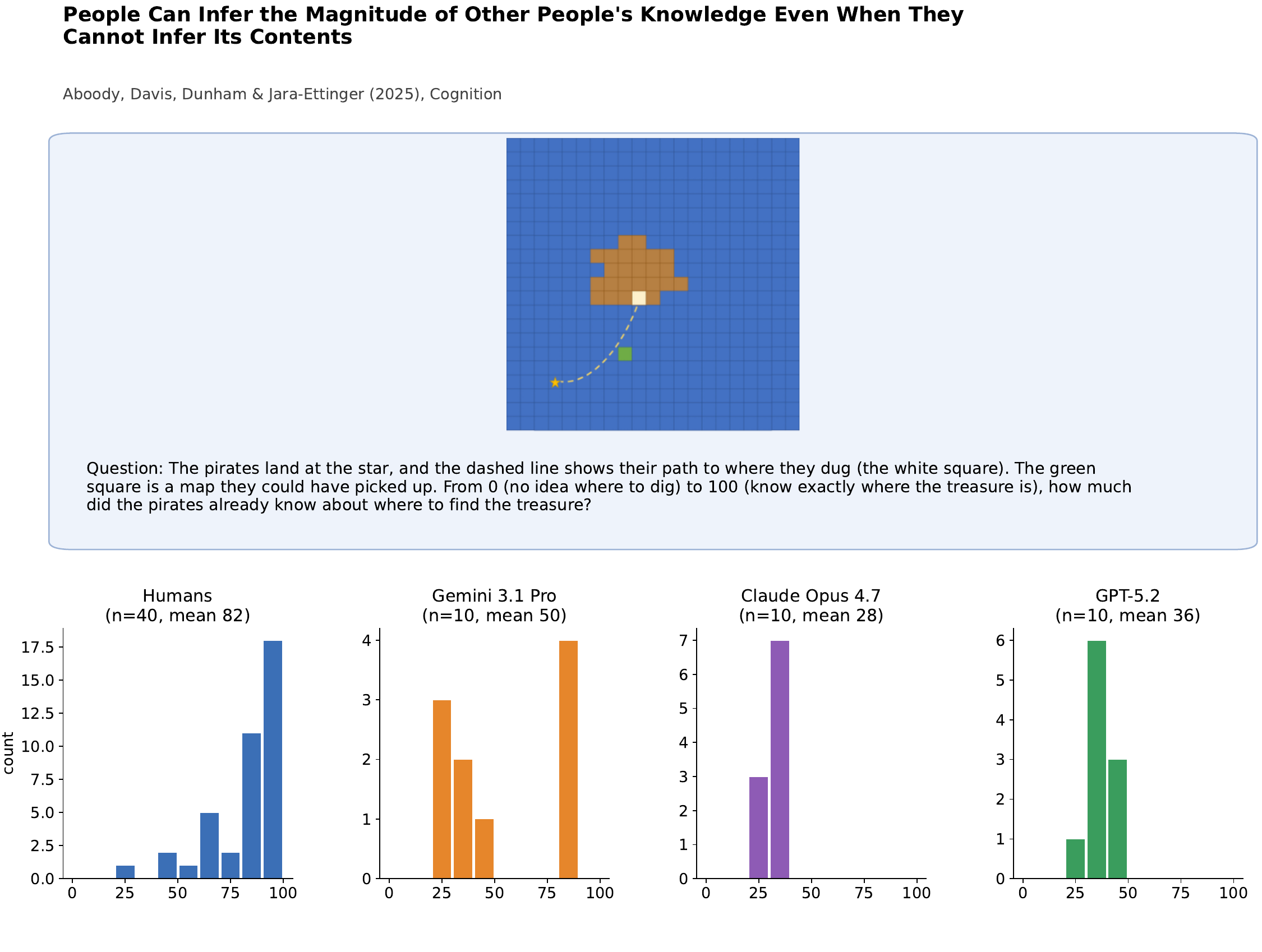}
    \caption{\textbf{Theory of mind (image): inferring knowledge from a skipped opportunity} \citep{aboody2025people}. Pirates land at the star and walk directly to where they dig (white square), passing close to a map (green square) without picking it up. Participants infer that the pirates already knew where the treasure was ($M=82$ on a 0--100 scale, $n=40$). Opus and GPT-5.2 answer $28$ and $36$ on average, and GPT-5.2 misdescribes the scene (\emph{\textquotedblleft{}They chose to detour to get the map, suggesting limited prior knowledge\textquotedblright{}}); Gemini splits between runs that read the scene correctly (\emph{\textquotedblleft{}The pirates skipped the map even though it was very close to their path, suggesting they already knew where the treasure was\textquotedblright{}}, answering $80$--$85$) and runs that answer $20$--$30$.}
    \label{fig:failure_aboody}
\end{figure}

\end{document}